\documentclass[article,12pt]{elsarticle}
\usepackage{placeins}
\usepackage{wrapfig}
\usepackage{float}
\usepackage{pgfplots}
\pgfplotsset{compat=1.17}
\usepackage{hyperref}
\usepackage{longtable}
\usepackage{pgf-pie} 
\usepackage{pifont}
\usepackage{wrapfig}
\definecolor{mygreen}{HTML}{3F6600}
 \usepackage{multirow}
 \usepackage{graphicx}
 \usepackage{lscape}
\usepackage{graphicx}
\usepackage{pgfplots}
\usepackage{caption}
\usepackage{bchart}
\usepackage{subcaption}
\usepackage{geometry}

\usepackage{graphicx} 
\usepackage{multirow} 
\usepackage{url}
\usepackage{footnote}
\definecolor{color0}{RGB}{102,194,165} 
\definecolor{color1}{RGB}{252,141,98} 
\usepackage{natbib}
\usepackage{graphicx}%
\usepackage{multirow}%
\usepackage{amsmath,amssymb,amsfonts}%
\usepackage{amsthm}%
\usepackage{mathrsfs}%
\usepackage[title]{appendix}%
\usepackage{xcolor}%
\usepackage{textcomp}%
\usepackage{manyfoot}%
\usepackage{booktabs}%
\usepackage{algorithm}%
\usepackage{algorithmicx}%
\usepackage{algpseudocode}%
\usepackage{listings}%
\usepackage{tablefootnote} 
\usepackage{comment}
\usepackage{hyperref}
\usepackage{amssymb}
\usepackage{longtable}
\usepackage{pgf-pie} 
\usepackage{graphicx}
\usepackage{pifont}
\usepackage{wrapfig}
 \usepackage{multirow}
 \usepackage{graphicx}
 \usepackage{lscape}
\usepackage{graphicx}
\usepackage{colortbl} 
\usepackage[normalem]{ulem}
\useunder{\uline}{\ul}{}

\usepackage{caption}
\usepackage{bchart}
\usepackage{subcaption}
\usepackage{pifont}
\usepackage{newunicodechar}
\newunicodechar{✓}{\ding{51}}
\newunicodechar{✗}{\ding{55}}
\IfFileExists{hideAllComments}{
  \def\hideAllComments{yes, to hide}
}{}
\ifx\hideAllComments\undefined
  \newcommand {\XXX}[1] {\textcolor{red}{XXX \bf #1}}
   \newcommand {\eutemplate}[1] {\textcolor{blue}{\textsl{#1} } }
   \newcommand {\discuss}[1] {\textcolor{purple}{\em #1}}  
   
\else
  \newcommand {\XXX}[1] {}
  \newcommand {\eutemplate}[1] {}
  \def\discuss#1{}
  
  \excludeversion{formsAbits}
\fi

\usepackage{hyperref}
\usepackage{amssymb}
\usepackage{longtable}
\usepackage{pgf-pie} 
\usepackage{pifont}
\usepackage{wrapfig}
 \usepackage{multirow}
 \usepackage{graphicx}
 \usepackage{lscape}
\usepackage{graphicx}
\usepackage{colortbl} 
\usepackage[normalem]{ulem}
\useunder{\uline}{\ul}{}

\usepackage{caption}

\usepackage{amssymb}

\begin{document}

\begin{frontmatter}



\title{Speech Recognition Transformers: Topological-lingualism Perspective}



\author[label1]{Shruti Singh}
\author[label2]{Muskaan Singh}
\author[label1]{Virender Kadyan}


\affiliation[label1]{organization={SoCS,University of Petroleum and Energy Studies},
            city={Dehradun},
            state={Uttarakhand},
            country={India}}
            
\affiliation[label2]{organization={SCEIS, Ulster University},
            addressline={Northland Road},
            city={Londonderry},
            country={UK}}            



\begin{abstract}


Transformers have evolved with great success in various artificial intelligence tasks.  Thanks to our recent prevalence of self-attention mechanisms, which capture long-term dependency, phenomenal outcomes in speech processing and recognition tasks have been produced. 
The paper presents a comprehensive survey of transformer techniques oriented in speech modality. The main contents of this survey include (1) background of traditional ASR, end-to-end transformer ecosystem, and speech transformers (2) foundational models in a speech via lingualism paradigm, i.e., monolingual, bilingual, multilingual, and cross-lingual (3) dataset and languages, acoustic features, architecture, decoding, and evaluation metric from a specific topological lingualism perspective (4) popular speech transformer toolkit for building end-to-end ASR systems. Finally, highlight the discussion of open challenges and potential research directions for the community to conduct further research in this domain.
\end{abstract}



\begin{keyword}
Automatic Speech Recognition, Review, Transformer, Monolingual, Bilingual, Multilingual, Cross-lingual, Large language Models


\end{keyword}

\end{frontmatter}



\section{Introduction}
With the emergence of transformer\cite{vaswani2017attention}, a lot of automation and transformation across the entire Artificial Intelligence (AI) landscape has taken place, especially with the wide adoption of Large Language Models (LLMs). The foundational goal of AI is to replicate human senses such as sight, hearing, touch, and smell. In a wider context, these modalities are a mode of communication tied to a particular sensory channel, like vision for the visual modality and hearing for the auditory modality \cite{sahay2020low}. \textit{For example, a sound recording scenario that captures the \textbf{conversation between two people} contains information about utterances of two distinct speakers, acoustic information; sound, noise, vibrations, linguistic content, and prosodic cues}. Fundamentally, an audio-based AI system must take in the audio, comprehend, and make inferences about the complex auditory information sources to achieve human-level accuracy. Some popular or key audio/speech-related tasks such as automatic speech recognition (ASR), question answering, speech enhancement, speech emotion recognition, and speech separation have gained significant interest.

Traditional ASR systems had inherent limitations due to their reliance on manually engineered features to extract linguistic content from speech signals, lacking a deeper understanding of language. This rigid and finite approach makes it difficult to adapt to language diversity. To overcome this, recent advances in end-to-end neural network architectures map directly from speech to text. It can efficiently transfer knowledge and learning to different languages and dialects. The key factor in the rising popularity of transformer architectures is the ability to capture long-term dependency compared to other deep learning models, such as recurring neural networks (RNN), which tend to encounter vanishing or expanding gradients. The transformer's success in speech and audio processing is the introduction of self- and multi-head attention mechanisms. It also allows parallel self-attentions across multiple attention heads, enabling the model to capture long input with contextual information\cite{voita2019analyzing}. With this paper, we urge you to aid the research on building ASR systems with foundational models for diverse languages.


\subsection{Scope and Taxonomy}
This paper provides a generic ASR system review that would be leveraged for all the downstream tasks such as (i) identifying cognitive impairments and dementia, (ii) investigating speech coding for data transmission, (iii) detecting hate speech, (iv) spoken content summarization or speech summarization (v) techniques for speech signal separation in a complex environment (vi) emotion recognition in speech (vii) identifying positive and hopeful speech (viii) facial animation and lip-syncing(ix) integration in multimodal and multi-resolution context  (x) speaker embedding and extraction (xi) advancement in voice activity detection (xii) usage in speaker diarization (xiii) multi speech context. Note that we do not cover the research studies that use speech recognition as one module for another speech task. Rather, we aim to focus on cutting-edge end-to-end speech recognition systems.
This paper implements a two-tier structured taxonomy in end-to-end automatic speech recognition to enhance readability and interdisciplinary integration. This taxonomy is organized around the dual dimensions of application and challenges. Firstly, it enables researchers to identify relevant applications in their domain, facilitating a deeper connection to related areas. Secondly, it allows for a unified, formula-driven comparison of model designs across different ASR domains, promoting cross-disciplinary collaboration. This approach provides a comprehensive view of ASR research, combining specific application insights with broader theoretical frameworks, thereby encouraging the breakdown of domain boundaries and fostering more effective communication and idea exchange.
\subsection{Related Surveys and Selection Criteria for Our Review Paper}
This paper correlates with previous surveys that have examined related work on ASR, specifically on the lingual aspect using state-of-the-art transformers. Earlier work \cite{dua2023noise} most prominently covered feature-engineering and machine-learning work. Recent surveys cover transformers and their modalities or application in vision, time-series, video, and speech or audio processing \cite{mehrish2023review, li2022recent}. To the best of our knowledge, we are the first to specifically highlight the development of speech recognition systems using state-of-the-art end-to-end transformer architectures and their self-attention mechanisms across linguistic approaches.
In this paper, We conducted a targeted literature review using Google    Scholar, focusing on recent academic publications \textit{within the last five years} that cover research on speech recognition using a transformer in a linguistic context. Our refined search included terms like \textit{Speech Recognition and Transformer} for \textit{Bilingual}, \textit{Multilingual}, \textit{Crosslingual} system. From the initial selection of 123 papers, we discarded 14 due to limited relevance or contribution. The remaining papers were categorized based on their focus: 32 on monolingual, 25 on bilingual, 28 on multilingual, and 24 on crosslingual speech recognition. This categorization aids in understanding research advancements in each linguistic category and highlights the most utilized corpora, languages, acoustic features, and models in these categories. This revision maintains the essential details while providing a clearer and more concise overview of the review process and its outcomes.

\subsection{Features and Contribution}
The paper has been structured to explore the aspect of lingualism in ASR, particularly focused on transformer architecture. Specifically, we cover (i) the background of traditional ASR, end-end transformer ecosystem, and speech processing using transformers and (ii) a review of existing state-of-art work based on linguist paradigm, i.e., monolingual, bi-lingual, multilingual, and cross-lingual (iii) overview and comparison of popular datasets, assessing their key characteristics and effectiveness based on benchmark results (iv) we examine and compare acoustic features, models, language models, and evaluation metrics used in various ASR systems (v)  we present a survey of the toolkit used for building or training foundational models, focusing on linguistically grounded speech recognition approaches. (vi) Finally, we highlight key challenges in this research area and suggest potential directions for the community to conduct further research.

The urgency of our work will help researchers develop resources and advance modelling techniques, especially in low-resource settings.
The rest of the paper is structured as follows. Section 2 presents an overview of the foundational models in developing ASRs. Sections 3, 4, 5, and 6 investigate datasets, languages, acoustic characteristics, and architectures for monolingual, bilingual, multilingual, and cross-lingual ASR systems. Section 7 provides an overview of popular speech transformers for ASR. Section 8 discusses the challenges that impact ASR performance in all these approaches and outlines future research directions in ASR systems. Finally, in Section 9, we conclude the paper.

\section{Background}
\subsection{Traditional ASR: a Brief History and Milestones}
Speech recognition systems have undergone significant progress since the early 1950s. A pioneering effort by Audrey at Bell Labs\cite{davis1952automatic} was to distinguish between digits from the audio, followed by the advancement of the MIT Lincoln Lab\cite{forgie1959results} that was able to recognize ten phonemes of spoken utterances. Later, such speech recognition systems were built that could recognise whole words \cite{velichko1970automatic}. These systems were speaker-specific and posed a serious problem with the demand for a larger vocabulary size to develop an efficient ASR system. In the 1980s, non-speaker-specific Hidden Markov Models (HMMs) \cite{lee1988large} were introduced for modelling speech and the n-gram language models to handle large vocabulary. In the 1990s and early 2000s, the HMM-GMM model \cite{alam2022bengali} was the dominant framework for more realistic scenarios with conversational speech, phone calls, conferences, etc. In the 2000s, neural networks became significantly popular, leading researchers as in \cite{bourlard1994connectionist} to combine feedforward neural networks with HMMs. It was in the mid-2010s when deep learning techniques such as long-short-term memory (LSTM) and RNNs could capture the seq-2-modelling of ASR.  Despite significant improvement in the state-of-the-art, these models could not capture long-term dependency or context, which was further overcome by the introduction of transformers.

\subsection{Traditional ASR: Architecture}
The traditional ASR architecture consists of (i) \textit{front-end}, which involves pre-processing such as noise reduction, normalization, voice activity detection, and feature extraction, transforms the audio signal into a set of features (like Mel-frequency cepstral coefficients or MFCCs) that represent the phonetic content of the speech, making it easier for the model to process. (ii) \textit{back-end} encompasses acoustic modelling (maps the extracted features to phonemes or speech units), pronunciation modelling (translates phonemes into words, considering variations in speech), language modelling (predicts word sequences, providing context and improving recognition accuracy), and decoding (combines these models' outputs to produce the final text transcription).
Traditional ASR systems were based on HMMs and Gaussian Mixture Models (GMMs) \cite{r54, r55} for acoustic modelling. HMMs are effective in modelling time-series data like speech, and GMMs are used to represent the probability distribution of different speech features. However, they have limited contextual understanding and cannot model the non-linear relationship between speech signals and variability in speech. Even the inherited property of GMM, such as the assumption of every feature as independent, does not correlate with the speech signal. Their lack of computational efficiency, scalability, and adaptability makes it harder to train the system on large-size datasets, new speech patterns, accents, or noisy environments. Some of these drawbacks of traditional ASR were overcome by seq-2-seq \cite{r56, r57} models such as RNN, which was able to process the entire audio sequence and output in a single step, eliminating the need for separate acoustic, pronunciation, and language models. Their recurrent architecture limits the parallel computation, leading to longer training times. Additionally, they struggle to process long audio sequences due to difficulty capturing long-range dependency, which was partially resolved by introducing attention mechanisms and transformers as in sec \ref{sec:transformers}.

\subsection{End-to-End Transformers: A Generic Architecture}
\label{sec:transformers} 

\begin{figure*} [h!]

\centering
\tiny
\includegraphics[width=13cm, height=6cm] %
{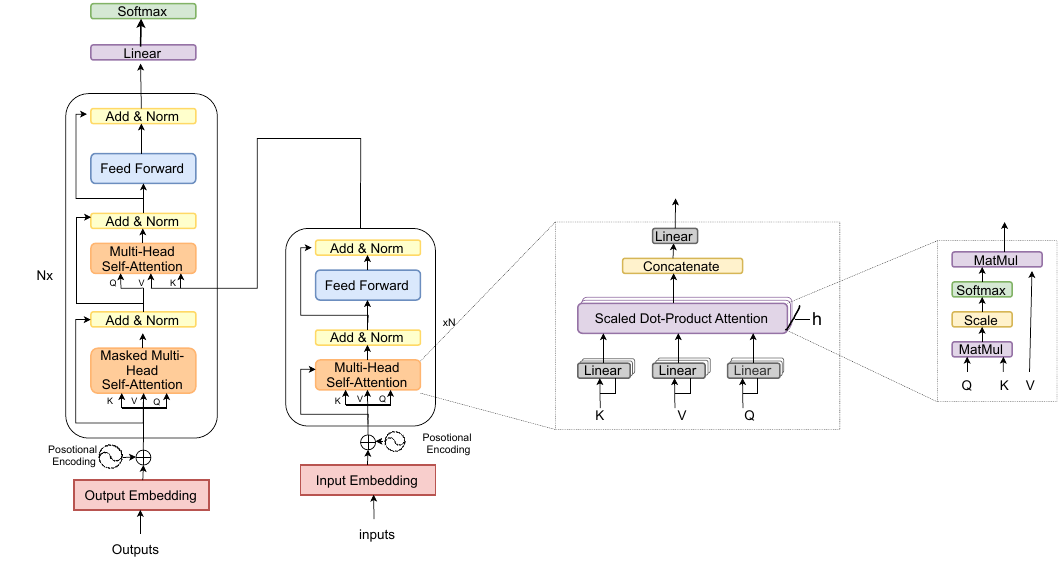}
\captionsetup{skip=0pt}
\captionsetup{justification=raggedright, singlelinecheck=false}
\caption{ (a) Audio/speech processing encoder-decoder architecture\cite{latif2023transformers}, \cite{vaswani2017attention},  with stacked self-attention and feed-forward sublayers. (b) with multi-head attention containing parallel attention layers executed simultaneously (c) Scaled dot-product attention}
\label{fig:transformer_architecture}       
\end{figure*}

As proposed by \cite{vaswani2017attention}, the vanilla transformer as end-to-end architecture is primarily introduced for machine translation and other sequence-to-sequence (seq-2-seq) tasks. This architecture represented a significant paradigm shift in natural language processing (NLP) tasks thanks to its innovative use of attention mechanisms. It is composed of 6 identical, stacked encoder-decoder blocks. These identical blocks contain multi-head self-attention (MHSA) and a position-wise fully connected feed-forward layer. It allows for longer relationships or dependency in data for large-scale language modelling. We briefly cover the architecture of the transformer with encoder-decoder blocks, specifically elaborating on self-attention and the multi-head attention mechanism. 
\textit{Encoder-Decoder Block} The encoder-decoder structure, as depicted in Fig. \ref{fig:transformer_architecture}, is further enhanced by the inclusion of a position-wise fully connected feed-forward network in each layer of both the encoder and the decoder. The encoder's main function is to process the input sequence \( x = (x_1, x_2, \ldots, x_n) \) into \( z = (z_1, z_2, \ldots, z_n) \). The decoder processes this continuous sequence of z to produce the output sequence \( y = (y_1, y_2, \ldots, y_m) \) elements.


\textit{Encoder Block}  Each encoder layer comprises two sub-layers: the multi-head self-attention (MHSA) layer and a position-wise fully connected feed-forward network. This feed-forward network consists of two linear transformations with a ReLU activation in between. The formula for the feed-forward network is \( \text{FFN}(x) = \max(0, xW_1 + b_1)W_2 + b_2 \). After each sub-layer, a process of layer normalization, expressed as LayerNormalization(x + Sublayer(x)), is applied.

\textit{Decoder Block} The decoder, like the encoder, is made up of a stack of identical layers, each comprising three sub-layers. These include a masked multi-head attention layer, a multi-head attention layer that attends to the output of the encoder, and a position-wise fully connected feed-forward network similar in structure to that of the encoder. The masked multi-head attention layer is unique to the decoder. It ensures that the predictions for a particular token depend only on the earlier tokens, not any future tokens. This is achieved by masking future positions within the self-attention layers of the decoder. After each sub-layer in the decoder, as in the encoder, layer normalization is also applied. Including the position-wise fully connected feed-forward network in the encoder and decoder is critical for handling variable-length sequences.

\textit{Embedding and Positional Encoding}
 In the Transformer architecture, the initial step involves tokenizing the input sequence, which segments the input into smaller, manageable units termed tokens. These tokens are then translated into embedding vectors, which serve as dense numerical representations of the tokenized input. A critical step in this process is the addition of positional encoding to these embedding vectors. This step is vital because Transformers process input in parallel and lack inherent mechanisms to discern the sequential order of tokens.
The positional encoding formula for each token position \( pos \) in the sequence is given by:
\[
PE_{(pos, 2i)} = \sin\left(\frac{pos}{10000^{2i/d_{\text{model}}}}\right), \quad PE_{(pos, 2i+1)} = \cos\left(\frac{pos}{10000^{2i/d_{\text{model}}}}\right)
\]
where \( pos \) denotes the position of the token in the sequence, \( i \) represents the dimension within the positional embedding, and \( d_{\text{model}} \) is the hidden dimension of the model. This method of positional encoding is essential in providing the model with information about the relative or absolute position of the tokens in the sequence, thereby enabling the Transformer to effectively understand and maintain the order of sequence elements during parallel processing.
\textit{Attention}
The attention mechanism is the most crucial component and is articulated through self-attention and multi-head attention modules.
\textit{Self-Attention Mechanism}: The self-attention mechanism updates each sequence element by aggregating global information from the entire input sequence. Mathematically, given an input \( Y \) in \( \mathbb{R}^{n \times d} \) space, where \( n \) represents the number of entities in the sequence and \( d \) is the embedding dimension, self-attention seeks to encode each entity using global contextual information. This is achieved through three learnable weight matrices, transforming the input \( Y \) into Queries (\( W^q \in \mathbb{R}^{d \times d_q} \)), Keys (\( W^k \in \mathbb{R}^{d \times d_k} \)), and Values (\( W^v \in \mathbb{R}^{d \times d_v} \)). The output \( Z \) of the self-attention layer, in \( \mathbb{R}^{n \times d_v} \), is computed as:
\[
Z = \text{softmax} \left( \frac{QK^T}{\sqrt{d_k}} \right)V
\]

Here, each entity's representation is a weighted sum of all entities in the sequence, with weights determined by the attention scores, which result from the query's dot product with all keys normalized by the softmax function.
\textit{Multi-Head Attention:}is designed to capture intricate relationships among different entities in the sequence. It comprises multiple parallel self-attention layers,  n each with its distinct set of learnable weight matrices (\( W^{Qi}, W^{Ki}, W^{Vi} \), for \( i = 0 \) to \( h-1 \), where \( h \) denotes the number of attention heads). For an input \( X \), the outputs of these \( h \) self-attention layers are concatenated into a single matrix, represented as:
\[
\text{MHA}(Q, K, V) = \left [ Z_{0}, Z_1, \ldots, Z_{h-1} \right ]
\]
This concatenated output is then projected onto a weight matrix \( W \in \mathbb{R}^{h \cdot d_v \times d} \). This multi-head approach allows the transformer to simultaneously process information from different representational subspaces at different positions, enhancing its ability to understand complex input sequences.
These mechanisms enable the model to capture long-range dependencies and understand the context within sequences, contributing to the model's effectiveness.

\subsection{End-to-End Transformer Architecture: for Speech Processing} 
For speech processing, transformers must handle continuous audio signals, which are longer and more complex than text. As described in section \ref{sec:transformers}, the encoder-decoder architecture is utilized with some alteration for speech processing as in Fig. \ref{fig:transformer_architecture}. The input audio is converted to suitable representation, often called spectrograms or mel spectrograms, to capture essential features of audio signals. The encoder processes these audio feature representations to capture the contextual or longer-term dependency in speech signals. The attention mechanism plays a critical role in speech-processing tasks such as speaker identification, emotion recognition, and speech translation, where understanding the broader context is crucial. They allow the model to focus on relevant parts of the speech signal, handling long-range dependencies.

Specifically, speech-based multi-head self-attention enables the model to simultaneously capture various aspects of the speech signal, such as phonetic content, speaker characteristics, and prosodic features. This processing capability is vital in dealing with speech's complex and varied nature. Additionally, positional encoding ensures that the model recognizes and preserves the order of the audio features. The output of this attention mechanism is then processed by a decoder that uses a feed-forward network to generate the next word or phoneme. Its output is further passed through the activation function, generating probability distribution over a vocabulary of possible phonemes or words for the next token in the sequence. This model employs different strategies to select the most probable token at each step during inference. The generated output is based on the application, such as textual transcriptions of the spoken content for speech recognition and audio waveform for speech synthesis. It effectively bridges the gap between between acoustic signals and linguistic content. Further, this encoding-decoding mechanism is coupled with pre-training and fine-tuning to adapt to diverse speech datasets or applications. These key modifications have led to significant advancement in various speech processing applications such as IVR, voice assistants, and mobile streaming ASR\cite{yeh2019transformer}, introducing novel models such as Emformer\cite{shi2021emformer}, Biencoder Transformer \cite{lu2020bi}, Squeezeformer \cite{kim2022squeezeformer}, Branchformer \cite{peng2022branchformer}, LFEformer \cite{wei2023lfeformer} or Speechformer \cite{chen2022speechformer} and modification as \cite{r118} achieved a notable result of 10.9\% WER on speech recognition benchmark dataset WSJ \cite{paul1992design}, with addition of convolution layers which downsamples long audio sequences in transformers.

\begin{figure}[!h]
\scriptsize
\begin{center}
\includegraphics [scale=0.5] {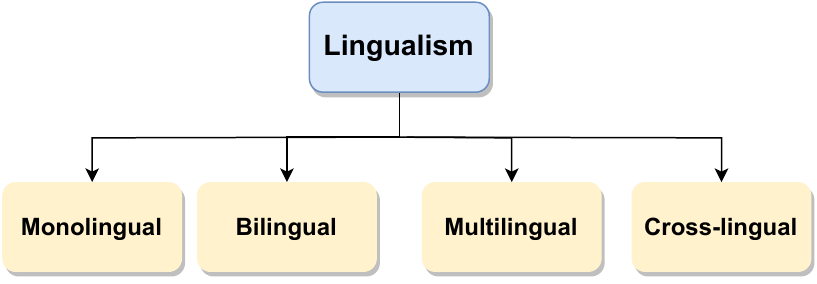} 
\caption{Audio processing taxonomy based on lingualism }
\label{lingualism_fig}       
\end{center}
\end{figure}

This paper focuses on advancing the research and innovation of ASR capabilities for diverse languages, specifically for low-resource or endangered languages.
The need for lingualism in ASR arises from bridging the communication gap, cross-cultural divergence, technology readiness, business growth, and development with advancement in linguistic,  sociolinguistic, and computational linguistic research. As the world becomes increasingly interconnected, the ability of ASR systems to handle multiple languages efficiently becomes more crucial.
To solve this, we propose a structured hierarchy to incorporate the aspect of lingualism in ASR, as depicted in Fig. \ref{lingualism_fig}. This hierarchy spans from monolingual to cross-lingual capabilities:(i) Monolingual ASR Systems: Understanding and processing speech in a single language. \textit{Discussion:} These systems are essential for precise language, accent, and dialect recognition. Due to the availability of extensive monolingual datasets, development and training are relatively straightforward, leading to significant advancements in voice assistants, dictionaries, and language-specific accessibility tools. Detailed explorations and research advancements are further discussed in Section \ref{sec:mono}. (ii) Bilingual ASR Systems: Capable of processing two distinct languages. \textit{Discussion:} Bilingual systems require the model to discern and process two languages, including their vocabularies, grammar, and pronunciations. A significant challenge is managing code-switching scenarios(inter-sentential and intra-sentential are missing) where speakers alternate between languages. Section \ref{sec:bi} elaborates more insights and research progress. (iii) Multilingual ASR Systems: Ability to understand and process multiple languages. \textit{Discussion}: These systems are crucial in diverse linguistic environments. The challenge lies in enabling the model to generalize across languages with varying phonetic, grammatical, and syntactical properties. Overcoming this often necessitates extensive and varied datasets or approaches like transfer learning. Additional details and research advancements are covered in Section \ref{sec:multi} (Low-resource perspective is missing). (iv) Cross-Lingual ASR Systems: Adaptation of knowledge from one language to aid in processing another, typically in data-scarce scenarios. \textit{Discussion:} Particularly vital for low-resource languages lacking extensive training data. Transfer learning techniques are commonly employed, where models trained on data-rich languages are adapted for other languages, aiding linguistic diversity and preserving lesser-spoken languages. Further discussions and advancements are detailed in Section \ref{sec:multi}\ref{sec:cross}.
From an ASR perspective, lingualism involves integrating multilingual capabilities into systems to enhance functionality. A monolingual ASR system should accurately transcribe speech from a single language. In contrast, bilingual or multilingual ASR systems must recognize and transcribe users' unique language, even when multiple languages are present in a single audio source \cite{r8}. Bilingual and multilingual speakers often exhibit intra-sentential and inter-sentential code-switching, shifting between languages within sentences or at discourse boundaries \cite{r12}. By integrating multiple languages and utilizing transfer learning, lingual ASR systems achieve improved speech recognition performance, effective code-switching handling, and enhanced accessibility for multilingual communities. Subsequent subsections will delve into recent advancements in monolingual, bilingual, multilingual, and cross-lingual ASR systems.

\section{Monolingual Models}
\label{sec:mono}
Recent trends have shown considerable progress in developing noise-robust ASR systems. Initially, a two-pass approach involving speech enhancement followed by speech recognition was common. However, researchers have observed that different noise types, such as non-stationary noise, require specific model-adaptation techniques for effective performance \cite{rownicka2020multi}. Semi-supervised learning (SSL) has also become crucial in ASR, especially for conformer models, which initially suffered from slow performance. Various optimization techniques have been proposed to enhance these models' efficacy, making them more suitable for real-world applications \cite{maison2023improving}.
A novel approach in modular end-to-end ASR systems combines acoustics-to-phoneme and phoneme-to-word networks, effectively addressing out-of-vocabulary (OOV) words \cite{r64}. This method demonstrates the potential of semi-supervised approaches in leveraging unlabeled data for ASR development.
Integrating transformers with CNNs, the conformer model emerged as a significant advancement in ASR \cite{gulati2020conformer}. Despite its initial slow performance, subsequent studies focused on various optimization techniques, significantly improving its efficiency and applicability in streaming applications \cite{peng2023i3d}.
Noise-robust ASR systems, essential for real-life scenarios, necessitate large training datasets \cite{r65}. Comprehensive comparisons of speech enhancement and model-adaptation techniques have been conducted, providing insights into improving noise robustness \cite{chen2023front}. Additionally, a gradient remedy approach has been adopted to enhance noise robustness in ASR systems further \cite{hu2023gradient}.
The lack of labelled data highlights the need for semi-supervised approaches in low-resource languages. Self-training methods in seq-to-seq models have substantially improved, utilizing large unlabeled datasets \cite{r26}.
Reinforcement learning has been explored as an effective method for speech enhancement, particularly using noisy-clean pair datasets \cite{r68}. This approach and a fully convolutional method for end-to-end systems signify a growing interest in innovative techniques to predict text from raw waveform data \cite{chen2022self}.
These advancements in monolingual ASR using deep learning, transformers, and reinforcement learning have overcome challenges in ASR, such as noise robustness and low-resource language processing. These developments depend on resources such as corpus, language, acoustic features, and novel architecture, leading to performance measures offering the system's efficiency to be usable. We further brief each of these in the section one by one.

\begin{table*}
\vspace{-10ex}
\scriptsize
\begin{tabular}{p{1.3cm}p{5cm}p{1cm}p{2cm}p{3.5cm}}

\hline
\textbf{Reference} & \textbf{Dataset} & \textbf{Language}  & \textbf{Duration} & \textbf{Performance} \\\hline
\cite{lu2020exploring} &
  Microsoft data &
  NA&65,000h & 
  12.2\% WER
   \\ 
  \cite{r28} &
  Librispeech &
  EN&960h &
  1.4/2.6\% WER\\ 
  \cite{gulati2020conformer} & 
  Librispeech &
  EN&970h &
  1.9\% WER\\ 
  \cite{r71} & 
  AISHELL &
  ZH&170h &
 10.04\% CER\\ 
 \cite{rownicka2020multi} & 
  Aurora-4, AMI &
  NA&21,100h &
(8.31,48.0)\% WER\\ 
 \cite{jain2020finnish} & 
  YLE news &
  FI&1500h &
17.71\% WER\\ 
\cite{Burchi2021} & 
  Librispeech &
  EN&960h &
  2.83\% WER\\ 
\cite{hsu2021hubert} & 
  Librispeech, Libri-light &
  EN&960h,60000h &
  1.8\% WER\\
\cite{chen2021developing} & 
  Microsoft data &
  NA&65000h &
  (7.69, 7.78)\% WER\\  
\cite{r65} & 
 RATS &
  EN&56h &
 60.7\% WER\\
 \cite{chen2022self} & 
 RATS &
  EN&44.3h &
 27.3\% WER, 17.6\% CER \\
  \cite{mehmood2022fednst} & 
Librispeech &
  EN&960h &
  3.31\% WER\\
 \cite{lehevcka2022exploring} & 
CommonVoice, VoxPopuli, MALACH &
  CS&111h &
  (5.75 9.01 12.93)\% WER\\
\cite{de2023boosting} & 
Norwegian Parliamentary
Speech Corpus (NPSC) &
  NO&126h &
  (11.54, 9.87)\% WER\\
\cite{saon2023diagonal} & 
Switchboard English, Switchboard+Fisher, MALACH &
  EN&(300,2000,176)h &
 (8.9, 9.0)\% WER\\
  \cite{andrusenko2023uconv} & 
Librispeech &
  EN&960h &
  3.5\% WER\\
 \cite{hu2023gradient} & 
 RATS, CHiME-4 &
  EN&57h &
 (12.2,13.4)\% WER\\
 \cite{chen2023front} & 
Gigaspeech, Librispeech &
  EN&(100,100)h &
  (19.8,19.3,18.1,14.13,13.79, 10.13)\% WER\\
\cite{maison2023improving} & 
 Common Voice, African Accented French, Canadian French Emotional, Corpus of
French as Spoken in Brussels &
  FR&(56.11,13.2, 1.09,4.07)h &
 (9.37, 5.29, 36.76)\% WER\\
 \cite{mai2023hyperconformer} & 
Librispeech &
  EN&960h &
  2.9\% WER\\
 \cite{peng2023i3d} & 
Librispeech &
  EN&960h &
  8.2\% WER\\  
 \cite{winata2020lightweight} & 
AiShell-1, HKUST &
  ZH&(150,5)h &
  (28.95, 13.09)\% CER\\  
 \cite{gao2022paraformer} & 
AiShell-1, AISHELL-2 &
  ZH&(178,1000)h &
  (5.2, 6.19)\% CER\\
\cite{chen2022wavlm} & 
Librispeech &
  EN&960h &
  1.8\% WER\\
\cite{baevski2020wav2vec} & 
Librispeech, LibriVox &
  EN&(960,60k)h &
  2.3,2\% WER\\
 \cite{xu2021transformer} & 
AiShell-1 &
  ZH&170h &
  6.49\%\\
\cite{kim2022squeezeformer} & 
Librispeech &
  EN&960h &
  2.47/5.97\% WER\\
\cite{shi2022streaming} & 
Voice commands, in-house TTS &
  EN&(13K,1.3K,4000)h &
  (16.22 4.32 6.12)\% WERR\\
 \cite{tian2020synchronous} & 
AiShell-1 &
  ZH&150h &
  8.91\% CER\\

\cite{maiti2024voxtlm} &  LibriSpeech
 & EN & 960h &
 3.5\% WER \\ 
\cite{tsunoo2024decoder} &
 LibriSpeech, Switchboard & EN & 960h, 300h & (3.2,8.7)\% WER \\
  \hline
  
 \end{tabular}
\caption{Audio/speech datasets for building/training a monolingual ASR system. Most of these datasets are in English with a duration range of  5-65k hours and achieve a significant CER/WER rate on test sets.}
\label{tab:monolingual_dataset_table}
 \end{table*}


\begin{table*}[t]
\vspace*{-1cm}

\scriptsize

\begin{tabular}[c]{p{1.5cm}p{1cm}p{3cm}p{5cm}p{1cm}}

\hline
\textbf{Reference} &
 \textbf{Noisy} &
 \textbf{Acoustic Features} &

\textbf{Architecture} &
\textbf{CTC}
\\
 \hline
\cite{lu2020exploring} &
\ding{55} &
80D-Mel filterbank &
Transformer-XL &
\ding{55} \\ 

\cite{r28} &
\ding{55} &
80D-Filterbank energy &
Conformer &
\ding{55}\\ 

\cite{r71} &
\checkmark &
Filterbank energy &

Transformer GRF &
\ding{55} \\ 

\cite{rownicka2020multi} &
\checkmark &
40D-Filterbank energy &

CNN &
\ding{55}\\ 

  \cite{winata2020lightweight} &
 \ding{55}  &
  NA  &
Standard transformer &
\ding{55} \\ 

\cite{jain2020finnish} &
\ding{55} &
NA &

Transformer-XL &
\ding{55} \\ 

\cite{baevski2020wav2vec} &
\ding{55} &
NA &

Transformer &
\checkmark \\

\cite{tian2020synchronous} &
\ding{55} &
40D Mel-filter bank &

Standard Transformer &
\ding{55} \\ 
 
\cite{gulati2020conformer} &
\checkmark &
80D-Filterbank feature &

Conformer &
\ding{55} \\ 

 \cite{hsu2021hubert} &
\ding{55} &
39D-MFCC &

Transformer &
\checkmark \\ 

\cite{Burchi2021} &
\checkmark &
80D-Filterbank energy  &

Conformer &
\ding{55} \\ 

  \cite{chen2021developing} &
\ding{55} &
 NA &

Conformer Transducer, Transformer Transducer &
\ding{55} \\

\cite{xu2021transformer} &
\ding{55} &
40D-Mel filter bank &

Standard transformer &
\ding{55} \\ 

\cite{chen2022self} &
 \checkmark &
 80D-Mel features & 
 
Conformer, Transformer &
\checkmark \\ 

 \cite{mehmood2022fednst} &
\checkmark &
NA &

Conformer &
\ding{55}
\\ 

\cite{gao2022paraformer} &
\ding{55} &
NA &

Standard transformer &
\ding{55} \\ 

\cite{chen2022wavlm} &
\ding{55} &
40D-Filterbank & 
Adapted transformer for the noisy environment &
\ding{55} \\
\cite{kim2022squeezeformer} &
\checkmark &
NA &

Transformer &
\checkmark \\ 

\cite{shi2022streaming} &
\checkmark &
80D Mel-filter bank &

Transformer transducer &
\ding{55} \\

\cite{radfar2022convrnn} &
\checkmark &
NA &
RNN-T  &
\ding{55} \\ 

\cite{r65} &
\checkmark &
80D-Mel features &
Conformer  &
\ding{55} \\ 

\cite{de2023boosting} &
\checkmark &
NA &

Transformer &
\checkmark \\

\cite{saon2023diagonal} &
\ding{55} &
40D-Mel features &

Transformer &
\checkmark
 \\
 
\cite{andrusenko2023uconv} &
\checkmark &
80D-Filterbank energy  &

Conformer &
\checkmark \\ 

\cite{hu2023gradient} &
\ding{55} &
Filterbank energy  &
Adapted transformer for the noisy environment &
\ding{55}\\ 

\cite{zorilua2023effectiveness} &
\checkmark &
80D-Filterbank energy &
Hybrid CTC/attention &
\ding{55}
\\ 
\cite{chen2023front} &
\checkmark &
Filterbank energy &

Transformer &
\checkmark \\ 

\cite{maison2023improving} &
\ding{55} &
NA &
Transformer &
\checkmark \\
 
\cite{kim2023branchformer} &
\ding{55} &
80D-Mel features  &
Transformer &
\checkmark \\
\cite{mai2023hyperconformer} &
\ding{55} &
80D-Filterbank &

Conformer &
\checkmark
 \\
\cite{peng2023i3d} &
\ding{55} &
NA  &

Conformer &
\checkmark \\

\cite{maiti2024voxtlm} & \ding{55} & NA & decoder-only & \ding{55}  \\

\cite{tsunoo2024decoder} & \ding{55} & NA & conformer-transducer & \checkmark \\

\hline 
\end{tabular}
\captionsetup{singlelinecheck=true}
\caption{Acoustic Features feed into the training of the model architecture of the Monolingual ASR systems. We also highlight the CTC and the noisy speech signals if applicable in the research study. }

\label{tab:monolingual_acoustic_table}
 \end{table*}

\subsection{Datasets and Languages}
In the context of monolingual end-to-end transformer ASR systems, as detailed in Section \ref{sec:mono}, specific datasets have emerged as popular choices for ASR. The Librispeech dataset, being the largest publicly available English corpus, is used most frequently, featured in 29.2\% of the studies \cite{r28, gulati2020conformer, Burchi2021, hsu2021hubert, mehmood2022fednst, andrusenko2023uconv}. AISHELL, a prominent Mandarin dataset, follows with a usage rate of 12.5\% \cite{r71, winata2020lightweight, gao2022paraformer, xu2021transformer, tian2020synchronous}. Other less commonly used datasets include CommonVoice \cite{lehevcka2022exploring, maison2023improving}, RATS \cite{r65,chen2022self, hu2023gradient}, Microsoft \cite{lu2020exploring, chen2021developing}, MALACH \cite{lehevcka2022exploring, saon2023diagonal}, YLE news \cite{jain2020finnish}, CHiME-4 \cite{hu2023gradient}, and HKUST \cite{winata2020lightweight}. These datasets vary in size, ranging from 50 hours to 60,000 hours. Our research findings, including dataset utilization, are summarized in Table \ref{tab:monolingual_dataset_table} and graphically presented in Fig. \ref{fig:mono_datasets}. English is the predominant language in monolingual ASR research, accounting for 64.3\% of studies \cite{r28, gulati2020conformer, Burchi2021, hsu2021hubert, r65, mehmood2022fednst}. Mandarin follows with 17.9\% usage \cite{r71, winata2020lightweight, gao2022paraformer, xu2021transformer}. Other languages like Bengali, Czech \cite{lehevcka2022exploring}, Norwegian \cite{de2023boosting}, Finnish \cite{jain2020finnish}, and French \cite{maison2023improving} have comparatively limited representation in the literature, comprising about 3.6\% of usage. Fig. \ref{fig:mono_languages} represents a different language distribution, highlighting the need for more research focused on low-resource languages. Librispeech, with its extensive English dataset, has been instrumental in achieving some of the lowest Word Error Rates (WER) in ASR, around 2\% from our analysis. AISHELL is a key resource for Mandarin ASR, with various versions like AISHELL-1 and AISHELL-2 offering 170 and 2000 hours of data, respectively. To enhance noise robustness in ASR systems, researchers often incorporate noisy datasets \cite{r71, rownicka2020multi, r65}. They are additionally augmenting smaller datasets to prevent overfitting, primarily through techniques like SpecAugment \cite{zorilua2023effectiveness, saon2023diagonal, r28, Burchi2021, r71}, speed perturbation \cite{chen2023front, kim2023branchformer, shi2022streaming}, and length perturbation \cite{saon2023diagonal}, has proven effective in improving ASR performance. We urge, \textit{benchmarking of datasets} specifically to English and Mandarin, with a growing need call for low-resource language datasets with acoustic variability to enhance the robustness of ASR.

\begin{figure}[H]
\tiny
  \begin{minipage}{0.45\textwidth}
    \centering
     \resizebox{6.5cm}{4.2cm}{%
    \begin{tikzpicture} 
\pie[  
hide number,
    pos = {8.5,0},
    rotate = 45,
    text = legend
    ] {
    4.25/CHiME-4(4.2\%), 
    4.2/Switchboard(4.2\%),
    4.25/MALACH(4.2\%),
    4.25/Libri-light(4.2\%),
    4.2/Microsoft data(4.2\%),
    4.2/CommonVoice(4.2\%),
    6.3/RATS(6.3\%),
    26.8/Other(26.8\%),
    29.2/Librispeech(29.2\%),
    12.5/AISHELL(12.5\%) } 
\end{tikzpicture}%
}
\captionsetup{skip=0pt}
\captionsetup{justification=raggedright, singlelinecheck=false}
\caption{ Presents our analysis for the usage of datasets for building/training the monolingual ASR system}
\label{fig:mono_datasets}

  \end{minipage}
  \hfill
  \begin{minipage}{0.45\textwidth}
    \centering
    \resizebox{6.5cm}{4.2cm}{%
      \begin{tikzpicture} 
\pie[  
                hide number,
                pos = {0,0},
                rotate = 45,
                text = legend
                ] {
                3.7/French (3.7\%),
                3.7/Norwegian(3.7\%),
                3.7/Czech(3.7\%),
                3.7/Bengali(3.7\%),
                3.7/Finnish(3.7\%),
                18.5/Mandarin(18.5\%),
                67.7/English(66.7\%)
                }
      \end{tikzpicture}%
}
 \captionsetup{skip=0pt}
 \captionsetup{justification=raggedright, singlelinecheck=false}
\caption {Presents our analysis of the usage of languages for building/training the monolingual ASR system. Commonly explored languages in the development of monolingual ASR are English, French, Norwegian, Czech, Bengali, Finnish, and Mandarin}
    \label{fig:mono_languages}
  \end{minipage}
\end{figure}

\subsection{Acoustic Feature}
In monolingual ASR, extracting and representing acoustic features are pivotal steps in processing speech signals. These features capture the essential characteristics of speech, enabling acoustic models to perceive and interpret the audio data accurately. Our analysis, is summarized in Table \ref{tab:monolingual_acoustic_table}, indicates a dominant preference for filterbank features, utilized in 76\% of the research \cite{r28, Burchi2021, r71, rownicka2020multi, andrusenko2023uconv, hu2023gradient, zorilua2023effectiveness, chen2023front, gulati2020conformer}. The remaining 23\% of studies employ Mel-Frequency Cepstral Coefficients (MFCC) features \cite{r65, saon2023diagonal, chen2022self, hsu2021hubert}. These feature extractors typically operate in 40-dimensional and 80-dimensional spaces. Features are normalized using mean and variance normalization to improve the system robustness \cite{radfar2022convrnn, saon2023diagonal}. MFCC works well in clean speech environments. By dividing audio signals into small frame-based segments, MFCC captures crucial spectral properties. However, its performance tends to decrease in noisy conditions. Filterbank features are increasingly preferred due to their direct representation of energy within frequency bands, avoiding the additional processing steps inherent in MFCC extraction. Their simplicity and efficiency make them ideal for real-time processing and tasks requiring high computational efficiency. Recent trends indicate a growing preference for filter bank features over MFCCs in monolingual ASR systems.

\subsection{Architecture-model and Decoding}
Both deep learning and transformer architectures are utilized for monolingual ASR, as in Table \ref{tab:monolingual_acoustic_table}. Our comprehensive analysis in Fig. \ref{fig:architecture_model_mono} indicates that transformer architecture is predominant, accounting for 89.7\% of the studies, while end-to-end deep learning architectures are employed in 10.3\% of the cases, as cited in \cite{rownicka2020multi, radfar2022convrnn, zorilua2023effectiveness}. The transformer architecture variations include vanilla transformer \cite{winata2020lightweight, tian2020synchronous, xu2021transformer, gao2022paraformer}, conformer \cite{r28, gulati2020conformer, Burchi2021, r65, andrusenko2023uconv}, and transformer-XL \cite{lu2020exploring, jain2020finnish}. Additional, forms like the transformer transducer \cite{chen2021developing}, adaptive transformer for noisy environments \cite{hu2023gradient, chen2022wavlm}, and transformers with CTC loss \cite{hsu2021hubert, chen2022self, de2023boosting, saon2023diagonal, chen2023front, maison2023improving, kim2023branchformer} are also highlighted. Conformers paired with CTC loss are also a topic of interest \cite{peng2023i3d, mai2023hyperconformer}. In terms of acoustic feature dimensions, transformers are applied to 40- and 80-dimensional features, whereas conformers are specifically associated with 80-dimensional features. Conformers have demonstrated approximately 2\% WER on the standard 960-hour Librispeech dataset\cite{r28, gulati2020conformer}. Decoding involves the selection of the most likely word sequence from the acoustic input, guided by language model probabilities. In monolingual ASR, our survey identifies two primary decoding mechanisms: beam search and greedy search. Beam search decoding, noted for its effectiveness in finding more accurate word sequences, is employed in several studies \cite{lu2020exploring, r71, chen2022self, saon2023diagonal}. On the other hand, greedy search, which is simpler and faster but potentially less accurate, is utilized in other research works \cite{radfar2022convrnn, mehmood2022fednst}.


\begin{figure}
\scriptsize
  \begin{minipage}{0.45\textwidth}
    \centering
     \resizebox{8cm}{3.5cm}{%
    \begin{tikzpicture} 
\pie[  
        hide number,
            pos = {8.5,0},
            rotate = 45,
            text = legend
            ] {
            3.4/Hybrid CTC/Attention(3.4\%),
            6.9/Adapted transformer for noisy environment(6.9\%),
            10.3/Conformer with CTC(10.3\%),
            3.4/RNN-T(3.4\%),
            24.1/Transformer with CTC (24.1\%),
            6.9/Transformer-XL(6.9\%),
            17.2/Conformer(17.2\%),
            20.7/Standard transformer(20.7\%),
            3.4/CNN(3.4\%),
            3.4/Transformer Transducer(3.4\%)
           }
\end{tikzpicture}
}

\captionsetup{justification=raggedright, singlelinecheck=false}
\caption {An analysis of architecture commonly used in monolingual ASR}
\label{fig:architecture_model_mono}
  \end{minipage}
  \hfill
  \begin{minipage}{0.45\textwidth}
    \centering
    \resizebox{7cm}{4cm}{%
           \begin{tikzpicture} 
\pie[  
hide number,
    pos = {8.5,0},
    rotate = 45,
    text = legend
    ] {
    9.1/AISHELL(9.1\%),
    9.1/Switchboard(9.1\%),
    6.1/ASRU 2019 challenge(6.1\%), 
    6.1/CommonVoice(6.1\%),
    15.2/SEAME(15.2\%),
    6.1/Self-recorded(6.1\%),
    6.1/Mandarin-English CS(6.1\%),
    15.2/Librispeech(15.2\%),
    30.4/Others(30.0\%)
     } 

\end{tikzpicture}%
}
\captionsetup{skip=0pt}
\captionsetup{justification=raggedright, singlelinecheck=false}
\caption {Usage of various datasets reported in bilingual studies}
\label{fig:dataset_bi}
  \end{minipage}
\end{figure}

\subsection{Evaluation Metric}

The performance of monolingual ASR systems is predominantly evaluated using Word Error Rate (WER) and Character Error Rate (CER). WER is the more widely used metric, featured in 83.3\% of the surveyed studies \cite{lu2020exploring, chen2023front, kim2023branchformer, peng2023i3d, mai2023hyperconformer}. It calculates the proportion of words the ASR system incorrectly identifies against a standard reference transcription, calculating word-level accuracy. On the other hand, CER, assessing errors at the character level, is used in 16.7\% of the research \cite{r71, winata2020lightweight}. It evaluates the accuracy at a character level, calculating the difference in characters between the recognized and reference transcriptions. CER is relevant in languages where characters form the core basis of the language structure, such as in logographic languages, where character-level distinctions are critical for understanding.CER becomes crucial in languages with complex orthographies or where character-level accuracy significantly impacts comprehension. Some studies have incorporated both metrics to provide a comprehensive evaluation of ASR performance \cite{mehmood2022fednst}.

\section{Bi-lingual Models}
\label{sec:bi}
Recent advancements in bilingual ASR systems have focused extensively on addressing code-switching and code-mixing scenarios, which are prevalent in bilingual speech. A significant challenge in this area is the scarcity of dedicated code-switching datasets. Researchers have tackled this issue by either leveraging existing monolingual datasets \cite{r79,r37, r34, r78} or augmenting them to simulate code-switching scenarios \cite{r32}. However, fine-tuning models for code-switched speech often leads to degraded performance in monolingual ASR. A plausible solution could be the Learning Without Forgetting (LWF) framework, which helps maintain monolingual accuracy while improving code-switching performance \cite{r37}. In \cite{r79}, the study addresses the challenge of intra-sentential code-switching, which affects the performance of monolingual speech. The research aims to preserve monolingual accuracy while enhancing code-switching speech recognition. Another critical issue is the reliance on language-dependent resources, prompting the need for ASR systems that minimize handcrafted inputs. Transformers have been helpful in bilingual ASR, particularly for handling code-switching where language switching occurs inter-sententially \cite{r78, r33}. These models are favoured for their fast inference and parallel computation capabilities \cite{r38, r33, deng2022improving}. In \cite{r31}, end-to-end attention-based ASR systems modifications include incorporating language identity information, workpiece models and transfer learning to handle bilingual speech effectively. This approach is particularly relevant for voice-assistant devices that require compact ASR models. Additionally, \cite{yan2023towards} explores the Transformer-transducer model for leveraging code-switching text data, addressing data scarcity in bilingual ASR tasks \cite{yu2023code, r12, yan2023towards}. CTC-based models have been a popular choice for bilingual ASR \cite{r15}. However, recent studies have shifted towards more sophisticated models like the non-auto regressive CTC-attention model proposed in \cite{deng2022improving}, which offers improved recognition accuracy and faster decoding. Hybrid models that combine supervised DNNs with i-vector feature extraction and phonological analysis have also shown promising results in enhancing ASR performance \cite{jain2023phonological, r75}. Moreover, with the growing demand for voice-assistant-enabled devices, the need for streaming-based, memory-efficient ASR models has been addressed in \cite{patil2023streaming}. The Bi-lingual ASR system has significantly progressed with deep learning and transformer models. These models can handle code-switching scenarios, optimize models for fast inference, and can streamline systems.

\subsection{Datasets and Languages}
We review the recent studies in bi-lingual ASR in section \ref{sec:bi} and present the findings of dataset and languages in Table \ref{tab:bilingual_dataset_info}.From the analysis presented in \ref{fig:dataset_bi}, some prominent datasets are SEAME, Librispeech, and AISHELL. SEAME is a Mandarin-English code-switching dataset encompassing 192 hours of recordings from a gender-balanced group of speakers aged 19-33, stands out in its utility. It has been a choice in approximately 15.2\%  of bilingual ASR research, demonstrating its relevance in studies exploring code-switching dynamics \cite{r34, r81, r33, r38, yan2023towards}. Librispeech, an equally significant dataset, was also used in 15.2\% of the studies. Researchers have leveraged this dataset independently or in conjunction with other monolingual or bilingual datasets \cite{r75, r15, r81, r33, r38, thomas2022efficient}, highlighting its versatility. AISHELL, prominently monolingual dataset, AISHELL has been adapted in 9.1\% of studies by combining it with other datasets like Switchboard and Librispeech \cite{deng2022improving, peng2022branchformer, r87}. This approach underscores the adaptability of AISHELL for bilingual ASR tasks. Speech perturbation \cite{r75, r34, r33, r38, yan2023towards, peng2022branchformer} and  SpecAugment \cite{r33, r38, r87, r81, peng2022branchformer}, augmentation techniques are employed to increase the data size and diversity. The analysis of language pairs presented in Fig. \ref{fig:bi_languages} indicates that Mandarin-English and Hindi-English are the most prominently used datasets. The most extensively researched Mandarin-English language pair featured in 44.4\% of the studies \cite{r31, r15, r81, r32, r33}. The focus on this pair reflects the global linguistic trends and the complexities involved in Mandarin-English code-switching. Hindi-English language pair accounting for 7.4\% of the research \cite{r86}, this language pair is gaining attention, indicative of the growing interest in South Asian linguistic contexts within ASR research. Apart from this, other language pairs like French-Mandarin \cite{thomas2022efficient}, Arabic-English \cite{r78}, and German-English \cite{r75}, each with a usage rate of 3.7\%, are less common but signal an expanding horizon in bilingual ASR research. The analysis focuses on certain language pairs, particularly Mandarin-English is the most prevalent for bi-lingual ASR systems. The selection of datasets like SEAME and Librispeech shows effective suitability for studying complex linguistic interactions. However, the relative underrepresentation of other language pairs suggests an opportunity for future research to diversify the linguistic scope of ASR systems.

\begin{table*}

\scriptsize
\vspace{-20ex}
\begin{tabular}{p{1.2cm}p{3.8cm}p{2cm}p{2.3cm}p{3.5cm}}
\\
\hline
\textbf{Reference} & \textbf{Dataset} & \textbf{Language} & \textbf{Duration} & \textbf{Performance} \\\hline
\cite{r15} &
  Microsoft live Cortana application in China, US-English Cortana &
  ZH,EN,ZH-EN& (4000,3400,300)h & 
  (12.23, 28.01)\% WER
   \\ 
  \cite{r34} &
  SEAME &
  EN, ZH&NA &
  (54.4, 45.6)\% MER\\ 
  \cite{r12} &
  FAME, Frisian Broad, CGN-NL &
  FRR,NL& (11.5,125.5, 442.5)h &
  31.4\% WER\\ 
   \cite{r31} &
  Mandarin-English CS data &
  ZH,EN&(1000h) &
  6.49\% CER\\ 
  \cite{r86} &
  NA &
  HI,EN,HI-EN& (353,133,4.5)h &
  56.97\% WER\\
\cite{r37} &
  Self-recorded &
  TA,TE,GU& (212,170,241)h &
  (55.65, 43.6, 44.2)\% WER\\ 
\cite{r75} &
  GerTV, Librispeech, Commonvoice, Fisher, Switchboard&
  DS, EN& (1000, 3000)h &
  8.78\% WER\\
\cite{r81} &
  SEAME &
  ZH,EN,ZH-EN& (21.7,19.9,51.24)h &
  18.8,28.5\% TER\\
\cite{r78} &
  MGB-2, TED-LIUM3 &
  AR,EN&450,1200h &
  50.1\% WER\\ 
\cite{r32} &
  ASRU 2019 Code-Switching ASR Challenge &
  ZH,ZH-EN& (200,500)h &
  10.29\% WER\\
\cite{r33} &
  SEAME, LibriSpeech &
  ZH,EN,ZH-EN,EN& (22.11,27.27, 65.2,960)h &
  16.6\% TER\\  
\cite{r38} &
  SEAME &
  EN,ZH&110h &
  (26.6,23.4)\% MER\\   
\cite{r87} &
  AISHELL-2, Librispeech, Mandarin-English CS &
  ZH,EN,ZH-EN& (1k,1k,13)h &
  47.84\% MER\\
\cite{peng2022branchformer} &
  Aishell, Switchboard,Librispeech &
  ZH,EN,EN& (170,300,960)h &
  (4.43,7.8,2.4)\% CER\\
\cite{thomas2022efficient} &
  Librispeech, Common voice &
  EN,FR& (960,1000)h &
  (12.91,28.25)\% WER\\
\cite{deng2022improving} &
  AISHELL-1, Switchboard &
  ZH, EN&NA &
  (4.3,11.9)\% WER\\
\cite{r80} &
  NA &
  ZH,EN& (1000,1000)h &
  9.9,7\% CER\\  
\cite{r79} &
  ASRU 2019 shared tasks &
  EN,ZH&700,500h &
10.2\%MER,8.1\% CER,29.2\% WER \\
\cite{patil2023streaming} &
  Self-recorded &
  EN,Hi,EN-HI& (10k,8k,2.6k)h &
  (16.56,14.68,31.69)\%WER\\
\cite{yan2023towards} &
  SEAME &
  ZH, EN&NA &
  26.0\% MER\\

\cite{jain2023phonological} &
  MUCS 2021 &
  Hi, GU,MR, TA, TE& (95.05,40,93.89, 40,40)h &
  (60.6,21,43.2, 28.2,26.5)\% WER\\

  \cite{singh2024mecos} & Self-recorded & EN-MNI & 57h &   (54.39,54.62,33.17,57.4)\% WER  \\

\cite{wang2024tri} & LibriSpeech, WenetSpeech-M, TALCS & EN, ZH, EN-ZH & (960,1000,23.6)h & (6.17,17.34,6.82)\% TER \\

\hline

\end{tabular}
\captionsetup{singlelinecheck=true}
\caption{Audio/speech datasets for building/training a bilingual ASR system. Most of this dataset is paired with English. Most of the research explores Mandarin-English language pair duration range of 25-8k hours and achieves a significant CER/WER rate on test sets.}  


\label{tab:bilingual_dataset_info}
\end{table*}

\begin{figure}
\tiny
  \begin{minipage}{0.45\textwidth}
    \centering
     \resizebox{8cm}{5.2cm}{%
      \begin{tikzpicture} 
\pie[  
                hide number,
                pos = {0,0},
                rotate = 45,
                text = legend
                ] {
                3.7/Marathi (3.7\%),
                3.7/Hindi (3.7\%),
                3.7/French-English (3.7\%),
                3.7/Arabic-English (3.7\%),
                3.7/German-English (3.7\%),
                3.7/Filipino-English (3.7\%),
                3.7/Frisian (3.7\%),
                3.7/Dutch (3.7\%),
                7.4/Gujarati(7.4\%),
                7.4/Telugu(7.4\%),
                7.4/Tamil(7.4\%),
                7.4/Hindi-English(7.4\%),
                44.4/Mandarin-English(44.4\%)
                }
      
      \end{tikzpicture}%
}
 \captionsetup{justification=raggedright, singlelinecheck=false}
\caption {Represents the language pair/languages with the most prominent being Mandarin-English for bilingual ASR systems}
    \label{fig:bi_languages}
  \end{minipage}
  \hfill
  \begin{minipage}{0.45\textwidth}
    \centering
    \resizebox{7cm}{4.2cm}{%
      \begin{tikzpicture} 
\pie[  
                hide number,
                pos = {0,0},
                rotate = 45,
                text = legend
                ] {
                25.0/CTC-Attention(25\%),
                20.0/CTC(20\%),
                10.0/LAS(10\%),
                5.0/Conformer with CTC(5\%),
                5.0/RNN-T(5\%),
                 5.0/DNN(5\%),
                20.0/Transformer with CTC (20\%),
                10.0/Standard transformer(10\%)
                }
      
      \end{tikzpicture}%
}
 \captionsetup{justification=raggedright, singlelinecheck=false}
\caption {Represents the diverse architectures utilized for the bi-lingual ASR system}
    \label{fig:bi_architecture}

  \end{minipage}
\end{figure}

\subsection{Acoustic Features}
Our investigation into recent bilingual ASR research reveals various acoustic features employed across various studies. This analysis, detailed in Table \ref{tab:bilingual_acoustic_table}, provides an insightful overview of the acoustic features predominantly used in this field. The most predominant acoustic features are MFCC, Filterbank, Pitch, and i-vector. MFCC is emerging as the most extensively used acoustic feature. It is employed in 40.0\% of the bilingual ASR studies \cite{r75, r78, r87, r79, r12, r32}. Its widespread adoption can be attributed to its effectiveness in capturing the key spectral properties of speech. The filter bank is the second most utilized feature, found in 32.0\% of the research\cite{r15, r31, r33, r38, r87, jain2023phonological, peng2022branchformer}. This feature is known for its ability to provide a comprehensive representation of the speech signal, making it highly suitable for ASR tasks. Pitch and i-vector features, accounting for 12\% and 16\% of usage \cite{r79, r33, r75, r78}, in bilingual ASR. Pitch features are crucial in capturing the prosodic aspects of speech, while i-vectors are instrumental in speaker and language recognition tasks within ASR systems.

\begin{table*}
\vspace{-10ex}
\scriptsize

\begin{tabular}[c]{p{1.5cm}p{1cm}p{1cm}p{3cm}p{3.5cm}p{1cm} }
\hline
\textbf{Reference} &
\textbf{Noisey} &
\textbf{CS/CM} &
\textbf{Acoustic Features}  &

\textbf{Architecture} &
\textbf{CTC }\\
\hline
 \cite{r15} &
 \ding{55}  &
 CS &
   80D-filterbank &

  CTC &
  \ding{55} \\
   \cite{r34} &
 \ding{55}  &
 CS &
 NA   &

  Hybrid CTC/Attention &
  \checkmark \\
   \cite{r31} &
\ding{55}   &
CS &
 80-D mel-filterbank &
  LAS &
  \ding{55}   \\ 
   \cite{r12} &

 \ding{55}  &
    CS &
    40D-MFCC, i-vector &
CTC &
\checkmark \\
   \cite{r86} &

\checkmark  &
  CM &
    MFCC &

  DNN &
  \ding{55} \\ 

   \cite{r37} &
   \ding{55} &
    CS &
    NA   &

CTC, BiLSTM &
\checkmark \\ 
   \cite{r75} &
\ding{55}   &
NA &
40-D MFCC, i-vector extractor &

   DNN &
   \ding{55} \\ 
   \cite{r78} & 
 \ding{55}  &
 CS &
 40-D MFCC, i-vector &

standard Transformer &
\ding{55} \\ 
   \cite{r32} &
  \ding{55} &
     CS &
    MFCC &   
 CTC &
 \checkmark \\
   \cite{r33} &
  \checkmark &
  CS &
  83D-MFCC, Pitch &
Conformer &
\checkmark \\
   \cite{r38} &
\ding{55}  &
 CS &
 80D-MFCC, Pitch &

Transformer &
\checkmark \\
   \cite{r81} &
  \checkmark &
 CS  &
  80D-MFCC &

 Transformer &
 \checkmark \\
   \cite{r80} &
  \ding{55} &
NA   &
 NA  &
 LAS  &
 \ding{55} \\ 
   \cite{r79} &
 \ding{55}  &
    CS &
     83D-MFCC, Pitch &

  Hybris CTC/Attention &
  \checkmark \\
   \cite{peng2022branchformer} &
\checkmark &
NA &
  80D-Filterbank &

Transformer &
\ding{55} \\ 
   \cite{thomas2022efficient} &
  \ding{55} &
NA  &
  NA &

 Transformer &
 \checkmark \\ 
   \cite{deng2022improving} &
  \ding{55} &
 NA &
 NA   &

 CTC/attention  &
 \checkmark \\ 

   \cite{r87} &
 \ding{55}  &
CS&
  80D-MFCC &

Transformer &
\checkmark \\ 
   \cite{patil2023streaming} &
 \ding{55}  &
 CM &
   80D-Filterbank &
RNN-T &
\ding{55} \\
   \cite{yan2023towards} &
\ding{55}   &
  CS &
  NA   &
CTC &
\checkmark \\
   \cite{jain2023phonological} &
\ding{55}   &
NA &
40D-MFCC, 80D-Filterbank &

   Hybrid CTC/Attention &
   \checkmark \\ 

   \cite{singh2024mecos} & \ding{55} & CS & MFCC, i-vector & DNN–HMM, TDNN, hybrid TDNN–LSTM and hybrid CTC-attention & \checkmark    \\

\cite{wang2024tri} & \ding{55} & CS & Filterbank & tri-stage training two-pass & \checkmark \\
   \hline
\end{tabular}
\captionsetup{width=\linewidth}
\caption{acoustic features most prominent being MFCC, CS-code switching, CM-code mixing for training the model in architecture. }
\label{tab:bilingual_acoustic_table}
 \end{table*}

\subsection{Architecture-model and Decoding}
The bilingual ASR architectures are advanced from deep learning to recently proposed transformers as presented in Table \ref{tab:bilingual_acoustic_table}. From our analysis, depicted in Fig. \ref{fig:bi_architecture}, more than 50\% of the ASRs employed deep learning architectures, while transformers were used only in 45\% of them. Deep learning architecture includes DNN \cite{r75}, RNN-T \cite{yan2023towards}, CTC \cite{r15, r37, r12}, hybrid CTC-attention \cite{r34, jain2023phonological, deng2022improving, r79} and LAS \cite{r80} models. The transformer models includes are standard transformer \cite{r78, peng2022branchformer}, Transformer with CTC \cite{thomas2022efficient, r87, r81, r38} and Conformer with CTC \cite{r33}. The standard transformers are used 40 \cite{r78}, and 80-dimensional \cite{peng2022branchformer} acoustic features, whereas the CTC-based conformer and transformers are used 80-dimensional features along with pitch features \cite{r33, r38}. Beam search \cite{deng2022improving, patil2023streaming} and greedy search \cite{yan2023towards, r79} decoding mechanism is used.

\subsection{Evaluation Metric} 

Our analysis of recent bilingual ASR studies highlights the prominent metrics as WER, MER, CER, and Token Error Rate (TER). The WER is the most widely adopted metric in bilingual ASR research; WER is employed in over 50\% of the studies. It is attributed to its effectiveness in quantifying word recognition accuracy in speech, particularly in languages where the word is a significant unit of meaning. The MER is utilized in 20\% of the research; MER emerges as a vital metric, especially in scenarios where bilingual ASR systems handle languages with differing linguistic structures. The CER and TER metrics are used less frequently than MER. CER is particularly relevant for character-based languages like Mandarin, where the accuracy of character recognition is more indicative of ASR performance than word recognition. In contrast, TER provides an alternative measure focusing on the accuracy of token recognition, which can be crucial in mixed-language scenarios.

\section{Multi-lingual}
\label{sec:multi}
Multilingual ASR systems have seen substantial improvements by integrating various transformer-based models. Studies like \cite{r44, r93} have shown that adding LVCSR and MOE layers to transformers can significantly enhance performance. Furthermore, phoneme-based methods, which are crucial in ASR as an intermediary step and have demonstrated drastic performance improvements \cite{zhang2023uml, liu2023towards, r93, r44, r125, yusuyin2023investigation, zhao2022improving}. These methods have been effective across supervised and semi-supervised approaches due to the shared phonetic features among different languages. Additionally, the use of sequence-to-sequence, LSTM, and transliteration models at various stages of ASR (acoustic modelling, decoding, and language modelling) has also been explored \cite{r44, r93, zhang2023uml}. The dataset's size and linguistic information play a critical role in multilingual training \cite{r46}. For instance, \cite{r125} found that phonetically related languages perform better in multilingual ASR systems compared to phonetically unrelated languages. This insight is crucial for training models on diverse language sets.
Phoneme-based approaches have been at the forefront of multilingual ASR research due to their ability to leverage similarities across languages. Foundational speech models in these systems enhance performance significantly \cite{ritchie2022large, yusuyin2023investigation, zhao2022improving, r43}. Before transformer architecture, grapheme-to-phoneme conversion is performed to robustify the model \cite{r43, ritchie2022large}. The investigation of phone-based Byte Pair Encoding (PBPE) for various language families further strengthens these findings \cite{yusuyin2023investigation}.
Innovations like the \textit{transformer transducer based streamable ASR } model, as proposed in \cite{zhu2020multilingual}, are based on a self-attention architecture with language identification (LID) capabilities. These models address the challenges of varying input sizes, which can create an imbalance in multilingual ASR systems.
The bert-CTC transducer model, as discussed in \cite{higuchi2023bectra}, outperforms the simpler bert-CTC model. This advancement indicates a significant step forward in enhancing the efficiency and effectiveness of multilingual ASR systems.
The multilingual ASR is advancing rapidly with the adoption of transformer-based models, phoneme-based approaches, and architectures like the bert-CTC transducer. These developments are crucial in addressing the unique challenges posed by multilingual speech processing, particularly in enhancing model performance for low-resource languages and managing the complexity of diverse phonetic structures.

\begin{table*}
\vspace{-8ex}
\scriptsize
\begin{tabular}{p{0.6cm}p{3.3cm}p{3cm}p{3cm}p{3cm}}
\hline
\textbf{Refe-rence} & \textbf{Dataset} & \textbf{Source Language} & \textbf{Target Language} & \textbf{Duration} \\\hline
\cite{parcollet2023sumformer} &
  Librispeech, CommonVoice &
  EN & 
  Nl, It, Fr & 
  (960)-(40,300,730)h 
   \\
\cite{radford2023robust} &
  LibriSpeech, Artie,Common Voice, Fleurs, Tedlium,CHiM\-E6, VoxPopuli,CORAAL, AMI IHM, Switchboard, CallHome, WSJ,AMI SDM1 &
  EN & 96 languages&(117113, 438218)h 

   \\   
\cite{xiao2021adversarial} &
 Common Voice, Diversity11, Indo12, Indo9, IARPA BABEL  &
  Tr,Tt,Ta,Sv,Mn,Lv, Dv,Br,Ar,En,Pt,Ru, Fr,De,Cy,It,Ca,Sv &
  Ky,Et,Es,Nl,Kab & (93,119.5)- (19,30)h 
   \\ 
\cite{zhu2020multilingual} &
  Nordics family  &
X & Da-dk,Fi-fi,Nb-no, Nl-nl,Sv-se &
(3.7,4.9,8.2,3.8,9.3)Mh-(4.2,3,15,15,12)kh \\

\cite{r46} &
 Euronews corpus  &
 X & Ds,Fr,It,Ru,Tr,Es,En & 
 (45,45,45,45,45,45,45)h \\ 
 
\cite{r43} &
 Amazon  &
  X & en,fr,es,ja,zh & NA \\

\cite{r93} &
 Google’s voice search  & 
 X & HI, BN, TA, KA & 
 NA \\

\cite{r123} &
SpeechOcean’s low-resource dataset  & 
X & GU, TA, TE & (40,40,40)h \\

\cite{r96} &
Google’s Voice Search traffic  &
En-us, En-in,Es-us, Pt-br,Es-es, Ar-gulf,Ar-eg, Hi-in, Mr-in,Bn-bd, Zh-tw, Ru-ru,Tr-tr, Hu-hu, My-my &
 En, De,Nl, Fr,Es, It,Pt & 
(53.5k,27.1k,47.6k, 32.9k,23.5k,11.9k,11.9k, 32.3k,16.7k,16.5k,22.8k, 22.8k,22.1k,9.9k,7.6k)h \\

\cite{r44} &
GlobalPhone  &
  Vi,Pt,Ru,Es,Sw,Sv, Ta,Th,Hrv,Tr,Uk,Ar, Bg,Zh,Cs,Fr,De,Ha,Ja, Ko,Po,Amh,Tir,Orm,Wal &  
Amh, Orm, Tir,Wal &  NA \\

\cite{r47} &
NA  &
X & Ar-eg,Ar-x-gulf, Ar-x-levant,Ar-x-maghrebi,En-us,Fr-fr &
(3600, 37k, 37k)h \\

\cite{r99} &
Multilingual LibriSpeech  &
X &
En,De,Nl,Fr,Es, Pt,Po & (44.5,6k,6k,6k, 6k,6k,6k,100)h \\

\cite{r125} &
Globalphone, AMH2020, AMH2005  & 
X & Hau,Swa,Uig,Ukr,Hrv, Tur,Deu,Kor,Bul,Spa, Swe,Pol,Tha,Vie,Jpn, Rus,Tir,Por,Fra, Orm,Cmn,Ces,Wal  & 
  (44.5,6k, 6k,6k, 6k, 6k, 6k,100)h \\	
  
\cite{chen2023improving} &
FLEURS  &
X &  102 languages& (1.4k)h \\

\cite{soky2023domain} &
EC  &
  KM,EN,FR & KM &(2,1,1)h  \\
  
\cite{tjandra2023massively} &
Multilingual Librispeech  &
X &  70 language & 150000h  \\

\cite{zhang2023uml} &
Google AI Principles, Voice Search, YouTube  & 
X & 
Ar,Zh,De,En,Es,Hi, It,Ja,Pt,Ru & (6.4,5.1,3.8,15.7,52.8,14.7, 29.8,21.3,11.5,20.7,12.4)kh \\
  
\cite{yang2023learning} &
Multilingual Librispeech  &
X & EN,FR,IT,NL & (44.7k,1.1k,0.2k,1.6k)h \\

\cite{liu2023towards} &
LibriSpeech, Libri-Light, Multilingual LibriSpeech, CommonVoice, LJSpeech & 
X &
  En,Nl,Fr,De,It, Pt,Es,Ky,Tt & (960,53.2k,100,1.8,4.6,24)h \\

\cite{karimi2022deploying} &
NA &
  ro, en, pt,zh,ja, it,fr, es,de & ro &
  (11.7k,77k)h \\  
  
\cite{zhao2022improving} &
OpenASR21, BABEL, IARPA MATERIAL & 
X & 
am,zh,fa,tl,sw,jv,so, gug,ku,vi,kk,mn,ps,ta & 
(10,10,10,10,10,10,10,10, 10,10,10,10,10,10,10,10)h\\ 
  
\cite{gaur2022multilingual} &
Nordics & da,fi,nn,sv,nl,en & da,fi,nn,sv,nl,en &
(3.3,3.8,5.4,7.5,8.4,52)kh \\

  \cite{ritchie2022large} &
MCV, NCHLT & 
X &
ak,ha,ig,nr,nso,rw,ss, st,sw,tn,ts,ve,xh,yo,zu & (253.07,169.38,144.73,51.58, 53.41,1405.75,53,154.23, 572.65,53.57,52.66,53.15, 53.15,139.61,1011.04)h \\
  
  \cite{wang2022lamassu} &
 Librispeech, Common voice &
  EN, DE &
  EN,DE,ZH & (10k,10k)h \\

 \cite{lin2023lexical} &
AISHELL, CSJ,TEDLIUM &
X & ZH,JA,EN & NA \\

\cite{lehevcka2024comparative} & CommonVoice,MALACH & X &  CS,EN,DE & 150k h \\

\cite{https://doi.org/10.48550/arxiv.2402.17954} & Common voice, Fleurs, VoxPopuli & X & ar, ca, cs, de, en, es, fi,fr, hu, it, ja,nl, pt,ro,ru,sk, st,sw, yo & NA \\

\hline
\end{tabular}

\captionsetup{skip=0pt}
\caption{Audio/speech datasets and languages for building/training a multilingual ASR system. It contains source and target language pairs. X shows that either the source language is explicitly not defined or the same as the target language. Most of the work contains common voice with a duration range of 4-680k hours and achieves a significant CER/WER rate on test sets. The combination of language ranges from 3-102.}

\label{tab:multilingual_dataset_info}
\end{table*}

\subsection{Datasets and Languages}
In multilingual ASR systems, datasets vary in size, language composition, and domain, each contributing uniquely to developing robust ASR systems. Our analysis, summarized in Table \ref{tab:multilingual_dataset_info} and depicted in Fig. \ref{fig:multi_dataset}, highlights recent multilingual ASR studies' most commonly used datasets. 
The Commonvoice dataset, which has 38 languages, was most commonly utilized in 11.1\% of the research works (cited in \cite{parcollet2023sumformer, radford2023robust, xiao2021adversarial, yusuyin2023investigation}). Followed by the Multilingual Librispeech dataset (7.4\% usage, \cite{r99, yang2023learning}), Google's voice search dataset (5.6\% usage, \cite{r93, r96}), and the Nordics dataset (3.7\% usage, \cite{zhu2020multilingual}). Other datasets include BABEL (\cite{xiao2021adversarial}), Fluers, TED-LIUM (\cite{radford2023robust}), and a wide range of corpora such as CSJ, AISHELL, NCHLT, MCV, IARPA MATERIAL, OpenASR21, LJSpeech (\cite{liu2023towards}), and Globalphone (\cite{r44, r125}). These datasets vary in size, ranging from 120 hours to a massive 150k hours (\cite{tjandra2023massively}). However, data size variations often lead to imbalances, with languages having larger data volumes typically demonstrating better model performance (\cite{parcollet2023sumformer}). To address this challenge, noisy datasets are often incorporated into model training, aiding in developing noise-robust ASR systems (\cite{zhu2020multilingual, zhao2022improving}). Data augmentation techniques are also employed to expand dataset sizes, consequently enhancing model generalization (\cite{soky2023domain, tjandra2023massively, zhang2023uml}). This dataset encompasses a broad spectrum of languages and acoustic conditions, providing a rich training ground for models to learn and adapt. 

\begin{figure}
\tiny
  \begin{minipage}{0.45\textwidth}
    \centering
     \resizebox{8cm}{4.5cm}{%
    \begin{tikzpicture} 
\pie[  
hide number,
    pos = {8.5,0},
    rotate = 45,
    text = legend
    ] {
    7.4/Multilingual LibriSpeech(7.4\%), 
    3.7/Glpbalphone(3.7\%),
    11.1/CommonVoice(11.1\%),
    5.6/Google’s Voice Search(5.6\%),
    3.7/Nordics(3.7\%),
    9.3/Librispeech(9.3\%),
    3.7/Fleurs(3.7\%),
    3.7/Tedlium(3.7\%),
    49.4/Other(49.4\%),
    3.7/Babel(3.7\%) } 
\end{tikzpicture}
}
 \captionsetup{skip=0pt}
\caption{Usage of various datasets reported in multilingual studies }
\label{fig:multi_dataset}

  \end{minipage}
  \hfill
  \begin{minipage}{0.45\textwidth}
    \centering
\resizebox{8cm}{4.5cm}{%
     \begin{tikzpicture} 
\pie[  
hide number,
    pos = {8.5,0},
    rotate = 45,
    text = legend
    ] {
    7.4/ CTC-Attention(7.4\%),
    7.4/DNN(7.4\%),
    18.5/RNN-T(18.5\%),
    22.2/Transformer with CTC (22.2\%),
    3.7/Streaming transformer (3.7\%),
    11.1/Conformer(11.1\%),
    14.8/Standard transformer(14.8\%),
    3.7/BLSTM(3.7\%),
    3.7/CTC(3.7\%),
    3.7/CNN(3.7\%),
    3.7/Transformer Transducer(3.7\%) }
\end{tikzpicture}
}
 \captionsetup{skip=0pt}
\caption{List of architecture explored for the development of multilingual ASR}
\label{fig:multi_architecture}
  \end{minipage}
\end{figure}

    
\subsection{Acoustic Features}
 Our analysis sheds light on the predominant acoustic features employed in recent multilingual ASR research. MFCC is the most frequently used feature employed in 39\% of the studies \cite{r93, r123}. They are known for their ability to mimic the human ear's response to different frequencies, thereby offering an efficient way to represent speech in ASR systems. Filterbank features are used in 27.3\% of the research \cite{zhu2020multilingual, lin2023lexical}. These features comprehensively represent the frequency bands within the speech signal, making them a valuable asset in multilingual ASR systems. Mel Spectrogram, another key acoustic feature, is employed in 21.7\% of the studies \cite{gaur2022multilingual, ritchie2022large, tjandra2023massively}. This feature represents the short-term power spectrum of sound and is particularly effective in capturing the intricate variations in speech across different languages. Pitch features, which convey information about the speech's tone and intonation, are utilized in 8.7\% of the research \cite{lin2023lexical, xiao2021adversarial}. These features are particularly useful in distinguishing linguistic nuances across languages. Finally, i-vector features, accounting for 4.3\% of the usage \cite{r125}, are often combined with either filterbank or MFCC features \cite{r125, lin2023lexical, xiao2021adversarial}. I-vectors provide a compact representation of speaker and channel characteristics, enhancing the model's ability to adapt to different speakers and recording conditions in a multilingual context.

\subsection{Architecture-model and Decoding}
Multilingual ASR systems are explored in different deep learning and transformer architectures presented in Fig. \ref{fig:multi_architecture}. More than 50\% 
The transformer architectures are mainly standard transformer \cite{liu2023towards, r43}, transformer transducer \cite{tjandra2023massively}, conformer \cite{r96, zhang2023uml, ritchie2022large}, streaming transformer \cite{zhu2020multilingual}, CTC-Attention transformer \cite{zhao2022improving} and transformers with CTC  \cite{parcollet2023sumformer, soky2023domain, yusuyin2023investigation}. These transformers have used 80-dimensional acoustic features. A decoding beam search and greedy search are used.

\begin{table}
\scriptsize
\begin{tabular}[c]{p{0.7cm}p{0.8cm}p{3cm}p{2cm}p{0.7cm}p{4cm}}
\hline
\textbf{Refe-rence} &
 \textbf{Noisy} &
\textbf{Acoustic Features} &
\textbf{Architecture} &
\textbf{CTC} &
\textbf{Performance} \\
 \hline
  \cite{r46} &
   \ding{55} &
MFCC, tonal features &
CTC &
\checkmark &
22.3\% WER \\ 
 \cite{zhu2020multilingual} &
  \checkmark &
80D log-mel Filterbank &
Streaming Transformer &
\ding{55} &
(12.1,14,15.8,17.7,13.4)\% WER \\
 \cite{r43} &
  \ding{55}  &
NA  &
 Transformer &
\ding{55} &
4.03\%PER,16.14\%WER \\ 

\cite{r93} &
  \checkmark &
 80D MFCC &
  RNN-T &
  \ding{55} &
  (21.6,20.8,25.6,30.5)\% WER \\ 
  
  \cite{r123} &
   \ding{55} &
 40D MFCC &
  CNN &
  \ding{55} &
  (18.36,21.24,20.92)\% WER \\ 
  
\cite{r96} &
   \ding{55} &
 80D MFCC &
Conformer &
\ding{55} &
7.9\% WER \\

 \cite{xiao2021adversarial} &
   \ding{55} &
80D log mel Filterbank, pitch &
attention-CTC &
\checkmark &
(50.72,64.4,51.18,43.35,32.19, 48.56)\% WER \\

\cite{r44} &
   \ding{55} &
 NA  &
DNN &
\ding{55} &
(13.4,29.34,22.43,28.86, 19.91)\%WER \\

\cite{r47} &
 \ding{55}   &
80D MFCC &
RNN-T &
\ding{55} &
(12.1,6.1,13.4)\%WER \\ 

\cite{r99} &
  \ding{55}  &
Log-mel filter bank &
  RNN-T &
  \ding{55} &
 (6.5,4.1,9.5,5.2, 3.7,8.2,8,6.6)\% WER  \\
  
\cite{karimi2022deploying} &
  \checkmark &
 80D Mel spectrogram &
  BiLSTM &
  \ding{55} &
  NA\\ 
  
 \cite{zhao2022improving} &
  \checkmark &
  MFCC &
CTC/attention, transformer &
\checkmark &
NA \\

\cite{gaur2022multilingual} &
  \ding{55}  &
80D Mel spectrogram &
RNN-T &
\ding{55} &
(7.6,13.8,9.8,8.8,9.8,5.3)\% WER \\

\cite{ritchie2022large} &
   \ding{55} &
 128C Mel spectogram &
Conformer &
\ding{55} &
(26.5,29.7,35,6.7,8.5, 9.8,3.8,4.7,6.4,1.4, 2.3,9.9,7.1,45.6,13.1)\%WER \\

\cite{wang2022lamassu} &
  \ding{55}  &
  NA &
  Transformer &
  \checkmark \\ 
\cite{r125} &
  \ding{55}  &
40D MFCC, 100D i-vector &
Hybrid HMM–DNN  &
\ding{55} &
(8.2,6.41,19.09,3.61,17.19,10.9,10.81, 13.73,8.12,5.7,21.01,11.21,8.44,5.2, 4.7,16.65,20.99,16.95,15.3,19.7,32.74, 19.49,17.7,14.65,22.87)\% WER\\ 

\cite{chen2023improving} &
  \ding{55}  &
 NA  &
  Transformer &
  \checkmark &
  10.1\% CER,31.5\% MER  \\ 
\cite{soky2023domain} &
  \checkmark &
  NA &
Transformer  &
\checkmark &
1.74\% CER  \\ 

\cite{tjandra2023massively} &
  \checkmark &
 80D Mel spectrogram & 
  Transformer transducer &
 \ding{55} &
 9.5\% WER \\ 
 
\cite{zhang2023uml} &
   \ding{55} &
  128D Mel Filterbank &
Conformer &
\ding{55} &
  (11.8,14.3,6.9,8.3, 6.9,10.6,20.4,10.3, 16.4,8.7,14)\% WER \\

  \cite{yang2023learning} &
   \ding{55} &
 80D Mel spectrogram &
Streaming RNN-T &
\ding{55} &
 (12.74,11.59,17.79,17.12)\% WER \\ 

\cite{yusuyin2023investigation} &
   \ding{55} &
 80D Filterbank & 
Transformer &
\checkmark &
(11.8,10.4,16.7, 13.7,17.7,14.1)\% WER \\ 
\cite{liu2023towards} &
  \checkmark &
  MFCC &
Transformer  &
\ding{55} &
NA\\

 \cite{lin2023lexical} &
  \ding{55}  &
80D Filterbank, 3D Pitch &
  Transformer &
  \checkmark &
  (4.88,3.73,10.6)\% CER \\ 
 \cite{radford2023robust} &
  \checkmark &
  80D Mel spectrogram &
 Transformer &
 \ding{55} & 
  (6.2,9,4.4,4.5,25.5, 7.3,16.2,16.9,13.8, 17.6,3.9,36.4,5.2)\% WER\\ 

  \cite{parcollet2023sumformer} &
 \checkmark &
NA   &
 Transformer &
 \checkmark & 
  2.2,10.8,10.4,31.5\% WER \\ 

  \cite{lehevcka2024comparative} & \ding{55} & NA &  Transformers & \ding{55} &  (16.97,38.35,22.86)\% WER \\

\cite{https://doi.org/10.48550/arxiv.2402.17954} & \ding{55} & Pitch, intensity, speaking rate & Transformer & \ding{55} & (12.68,17.51,13.59,19.87, 16.42,12.74)\% WER \\

 \hline
\end{tabular}
\caption{The acoustic feature used for training model in the end-to-end architecture of Multilingual ASR systems}
\label{tab:multilingual_acoustic_info}
 \end{table}

\subsection{Evaluation Metric}
There are predominately four metrics used to evaluate multilingual ASR: WER, CER, Phoneme Error Rate (PER), and MER. WER is mostly used in Indo-European languages. Research shows that WER is a reliable metric for evaluating ASR performance in languages with a syntactic structure similar to English \cite{chen2023improving, soky2023domain}. CER For languages like Chinese and Japanese, where word delimitation is not space-based, CER becomes a more appropriate metric. Studies employing CER focus on the character-level accuracy of ASR systems, crucial for handling logographic scripts \cite{lin2023lexical}. This metric provides insights into the system's precision in transcribing each character, a critical aspect of accuracy in these languages. PER is used for phoneme recognition \cite{r43}, especially in languages with complex phonemic structures. It measures the system's capability to accurately identify and transcribe individual phonemes, which is particularly challenging in multilingual contexts with diverse phonetic systems. MER is employed in studies that tackle the challenge of language diversity within a single ASR system \cite{chen2023improving}. This metric combines WER, CER, and PER elements to provide a more holistic evaluation of ASR systems operating on languages with varying structural features.
\section{Cross-lingual}
\label{sec:cross}
Cross-lingual ASR systems have evolved significantly, addressing unique challenges through diverse training approaches and deep learning techniques. These advancements are particularly crucial in managing the complexities associated with cross-lingual speech recognition, especially in low-resource language settings. The \textit{training methodologies for cross-lingual ASR} systems span a broad spectrum from supervised to unsupervised methods. A prominent supervised approach involves phoneme-based training for knowledge transfer. Notable works in this area include the development of a phone inventory for unseen languages, leading to the creation of universal phone-to-articulatory mappings \cite{feng2021phonotactics}. These efforts aim to bridge the gap between languages and improve the system's ability to adapt to new linguistic contexts. Studies like \cite{wiesner2022injecting} suggest that extracting diverse \textit{learning representations from raw waveform} can significantly enhance knowledge transfer. This concept is further explored in works like \cite{likhomanenko2023unsupervised} and \cite{feng2020unsupervised}, which delve into the intricacies of unsupervised learning and its application in cross-lingual ASR. Recent advancements have also been made in \textit{data injection into phone language models}, especially in few-shot settings \cite{r127, wiesner2022injecting}, along with iterative \textit{pseudo-labeling} \cite{silovsky2023cross}, phone sequence decipherment \cite{klejch2021deciphering}, and cross-lingual self-training \cite{zhang2021xlst}. These methods aim to enhance the model's capability to handle a variety of linguistic inputs and improve accuracy in cross-lingual contexts. The introduction of\textit{ meta-learning methods}, as seen in \cite{r52}, addresses the issue of full-model adaptation, which is computationally intensive. The proposed meta-adapter offers a solution with faster computation and fewer parameters, making it suitable for \textit{cross-lingual adaptation}, particularly in low-resource scenarios. In the domain of unsupervised transformers, \cite{r103} emphasizes the use of\textit{ Contrastive Predictive Coding (CPC)}, utilizing pre-trained phoneme representations from a source language and applying them to target languages. This approach underscores the importance of leveraging existing linguistic knowledge to facilitate learning in new language environments. Deep learning plays a pivotal role in the development of cross-lingual ASR systems. Techniques like \textit{data injection}, \textit{iterative pseudo-labeling}, \textit{decipherment}, and \textit{knowledge distillation} \cite{fukuda2021knowledge, r127, klejch2021deciphering} have been crucial in overcoming the challenges associated with low-resource languages and enhancing the overall efficacy of ASR systems.
A critical challenge identified in cross-lingual ASR is \textit{overfitting}, especially in low-resource language contexts \cite{r49}. Addressing this issue is key to developing robust and adaptable ASR systems that efficiently handle a wide range of linguistic inputs. The field of cross-lingual ASR integrates advanced training methods, deep learning techniques, and unsupervised learning approaches to overcome the challenges.

\begin{table*}
\vspace{-8ex}
\scriptsize
\captionsetup{singlelinecheck=true}
\caption{Audio/speech datasets for building/training a crosslingual ASR system. The total languages in source and target are 79 and 57. English in the source and Spanish are the targets of most of the research work.}

\begin{tabular}{p{0.8cm}p{3cm}p{5.5cm}p{1.2cm}p{2.5cm}}
\hline
\textbf{Refe- rence} & \textbf{Dataset} & \textbf{Language} & \textbf{Duration} & \textbf{Performance} \\\hline

\cite{r50} & Librispeech, Thchs30 & En-Zh & 100-27.2hr & NA \\ 

\cite{r49} & Fisher, Babel & En-En & (150-30) & NA \\ 
\cite{r51} &
  Librispeech, Common Voice, MuST-C, Voice v42, IARPA Babel &
  (En, Fr)-(Zh,Pt,Nl,Mn,Vi,Ht) & &
    5.7\% WER \\   
 
\cite{r52} & Common Voice & (Chv, Mt,El, Lv,Cnh, Ky,Dv, Sl,Frr, Sah)-(Br, Sb,Ga, Ro,Ori) & & 64.1, 75.7, 67.0, 79.9, 64.3\% WER \\

\cite{r103} &
  Librispeech, Zerospeech, Common Voice &
  (En, Fr,Zh)-(Es,It,Ru,Tt,Fr) & &
  30.7\% PER\\

\cite{conneau2020unsupervised}   &
Common Voice, Babel, Librispeech &
(Fr, Nl, Ru, Sv, Ky, It, Tr, Bn, Ka, Zh, Ht, Tr, Ku, Ps,  Ta, Tpi, Pt, Pl, Nl, De, Fr, En, Vi, Tt, Es, Zh)-(As,Tl,Sw,La) 
& (55,76, 39,59)

 & 17.9, 13.1, 21.3, 22.4\% PER  \\

\cite{feng2020unsupervised} &
CGN,Aidatatang 200zh, Libri-light &
  (Nl,Zh)-(En,Nl,Zh) & (483,140)h & 
Nl:8.98\% WER, Zh:6.37\% CER
  
  \\ 

\cite{klejch2021deciphering} &
  Librispeech, GlobalPhone  &
(En, Fr,Ds, Es,Ru, Pl)- (Bg,Cs,Ha,Pt,Sw,Sv,Uk) &
(20,110)h  &
10.7,16.5, 30.3, 21.2, 98.3\% WER\\

\cite{r116} &

MLS, Multi-Genre Multi-Dialectal Broadcast News LibriLight, Librispeech &
  (En)-(Fr,Ar,De,It,Po,Es,Pt,Nl) &  &
  20.8\% WER\\
   
\cite{polak2021coarse} &
  LibriSpeech, Czech Parliament Plenary Hearings, CommonVoice &
En-Cs & 400h & 16.57\% WER \\
  
\cite{r127} & 

IIT Madras dataset, Google indic   &
  (Hi, Gu,Mr, Ta,Ml, Ka)-(Gu,Ka,Mr,Ml) &
  (183-25)h& 
  13,15,6,3\% PER\\
 
\cite{fukuda2021knowledge} &
  NA &
  (En, Ja,Ko, Pt-br)(En, Ds) & &
 19.2, 20.2\% WER
 \\
\cite{r53} &
Common Voice &
(Ru,Cy,It,Eu,Pt-Ro,Cs,Br,Ar,Uk)&
(355-52)h & 
47.29, 34.72, 59.14, 46.39, 47.41\% WER \\
\cite{ashihara2023exploration} &
CSJ, LTVS, JNAS, JSUT, TED50h &
(Ja,En-Ja,En)&
(2604)h &  
3.5, 13.0, 11.3, 13.9\% CER \\
\cite{r108} &
Multilingual Librispeech &
En-(De,Nl,Fr,Es,It,Pt,Po)&(44.6k-5.97k)h 
 & 8.4\%, 9.7\% WER\\

\cite{silovsky2023cross} &   
Naturally weighted Balanced &
(Cs,Pl,Ru,Sk,Uk)-Uk & (686-109)h  & 25\% WERR \\
\cite{likhomanenko2023unsupervised} &
Common Voice &
  (En,Ds,Es,Fr,Rw)-(En,Ds,Es,Er,Sw,Ha) && 15, 18\% WER\\
\cite{farooq2023learning} &
  IARPA BABEL &
(Ta,Te,Ceb,Jv)-(Ta,Te,Ceb,Jv)&(170)h & 
(38.83, 52.06, 29.94, 30.47)\% CER  \\
\cite{xue2023tranusr} &
Common Voice, TED2020   &
   En-(Es,Fr,It,Ky,Nl,Ru,Tt) & &
(5.5, 6.5, 7.1, 6.8, 8.0, 8.8, 7.2)\% PER  \\

\cite{hu2024cam} &  FLEURS, Common Voice & (BG,UK,BE,SL,SK,CS)- (BG,UK,BE,SL,SK,CS) & 65h & 23.3\% WER \\

\cite{chen2024improving}
& IARPA BABEL, OpenSLR, Common Voice & 

(BN, TL,ZU, LI, GN,TR,GU, Colombian ES, Chilean ES, TA, Peruvian ES, GL, KA, Northern EN, EU, EN, FR, IT, SW, DE, PT)-(VI, SW, Kurmanji, TA, YO, Argentinian ES, ML, MR, Nigerian EN, MY, Venezuelan ES, CA, CY, RU, ES, NL, Kabyle) & (477.95-298.93)h & (44.77, 44.37, 53.72)\%CER \\

   \\ \hline
\end{tabular}   


\label{tab:crosslingual_dataset_info}
\end{table*}

\subsection{Datasets and Languages}
The advancement of ASR in cross-lingual contexts heavily relies on diverse datasets that cover a broad spectrum of languages and dialects. This section delves into the significant datasets and languages employed in cross-lingual ASR research, highlighting the challenges and trends in dataset usage and language coverage. 
\textit{Commonvoice dataset} is a prominent resource in multilingual ASR research, offering a substantial collection of 2500 hours of speech data across 38 languages \cite{ardila2019common, r52, r53, r103, conneau2020unsupervised, r51}. Commonvoice is characterized by its extensive language diversity and substantial participant involvement, making it a valuable asset for cross-lingual studies. \textit{Librispeech} focused on English, this dataset comprises 1000 hours of read speech sampled at 16kHz, primarily sourced from public audio domains \cite{panayotov2015librispeech, r50, r51}. Its large volume and high-quality recordings are crucial for English-centric or multilingual studies involving English. \textit{IARPA Babel}, is towards conversational speech; the Babel dataset includes ten Asian and African languages. It provides audio recordings in real-life scenarios and diverse acoustic conditions. The dataset offers two configurations: the Full Language Pack (FLP) with 80 hours and the Limited Language Pack (LLP) with 10 hours of speech, making it suitable for resource-constrained language studies \cite{gales2014speech, conneau2020unsupervised, r49, r51, wiesner2022injecting}. Globalphone, a less common but significant dataset, covers 20 languages in 4000 hours, half of which are from native speakers. It has been a valuable source in cross-lingual research, especially for languages underrepresented in other datasets \cite{schultz2013globalphone, r50, r51, r116}. Japanese Corpus, consisting of 2500 hours of Japanese speech, including JSUT, LTVS, and JNAS, is one of the largest for a single language. It is complemented by additional datasets like CGN and Aidatatang 200zh, providing extensive resources for Japanese-centric ASR research \cite{ashihara2023exploration, feng2020unsupervised}. Our analysis in Table \ref{tab:crosslingual_dataset_info} shows that the total number of languages covered in source and target configurations are 79 and 57, respectively. English is predominantly a source language, while Spanish is a frequent target. This dataset size and language diversity imbalance often leads to model biases, particularly towards well-represented languages \cite{r127}. Recent research has explored various language pair combinations, including one-to-one \cite{r50, polak2021coarse}, one-to-many \cite{xue2023tranusr, r116}, and many-to-many \cite{klejch2021deciphering} configurations. These studies have shown significant improvements in ASR performance, especially for low-resource languages, demonstrating the field's growing capability to accommodate and enhance speech recognition across various linguistic contexts. The diverse range of datasets allows researchers to tackle challenges associated with low-resource languages, dialectal variations, and acoustic diversity.
\subsection{Acoustic Features}
Cross-lingual systems handle a diverse range of phonetic and prosodic characteristics. We review and summarize the usage and impact of various acoustic features in cross-lingual ASR tasks, as represented in Table \ref{tab:cross_acoustic}.
The most prominently used acoustic features in cross-lingual ASR are MFCC, Filterbank, Pitch, Bottleneck, and I-vector features. MFCC works well in clean speech environments \cite{r113, r50, r49, r51, feng2020unsupervised, feng2021phonotactics}. By dividing audio signals into small frame-based segments, MFCC captures crucial spectral properties. However, its performance tends to decrease in noisy conditions. Filterbank features are increasingly preferred due to their direct representation of energy within frequency bands, avoiding the additional processing steps inherent in MFCC extraction \cite{ashihara2023exploration, r108, farooq2023learning}. Their simplicity and efficiency make them ideal for real-time processing and tasks requiring high computational efficiency. Recent trends indicate a growing preference for filter bank features over MFCCs in cross-lingual ASR systems. Pitch features capture speech prosody and intonation, particularly the fundamental frequency (F0), essential in cross-lingual ASR \cite{r52, r53}. These features extract a time series of pitch values per frame, enhancing ASR accuracy by incorporating prosodic information, speaker characteristics, and effective segmentation. Bottleneck features are less commonly used and play a significant role in speaker identification and verification, utilizing the vocal characteristics unique to individuals \cite{r101}. I-vector features are crucial for speaker verification, offering a compact representation of speaker characteristics \cite{r103}. Their usage, however, is less frequent compared to other features. There has been a notable shift towards filterbank features in recent years due to diverse linguistic environments and computational resources. The increased adoption of filterbank features reflects the evolving landscape of ASR technology, where the focus is not just on accuracy but also on adaptability and efficiency, especially in multilingual and diverse acoustic scenarios.

\begin{figure}
\tiny
    \centering
     \resizebox{8.5cm}{4.5cm}{%
    \begin{tikzpicture} 
\pie[  
                hide number,
                pos = {0,0},
                rotate = 45,
                text = legend
                ] {
                 4.2 / Hybrid CTC-Attention(4.2\%),
                 8.3 / Hybrid DNN-HMM(8.3\%),
                 4.2/Transformer Transducer(4.2\%),
                 4.2/Standard transformer(4.2\%),
                 4.2/Conformer(4.2\%),
                 8.3/RNN-T(8.3\%),
                 8.3/CTC(8.3\%),
                 4.2/LSTM(4.2\%),
                 8.3/CNN(8.3\%),
                 12.5/DNN(12.5\%),
                 4.2/ Hybrid CTC-Attention transformer(4.2\%),
                 8.3/ Adapted Crosslingual Transformer(8.3\%),
                 4.2/ Crosslingual transformer(4.2\%),                
                16.7/Transformer with CTC (16.7\%)
                 }

            \end{tikzpicture}
            }
\caption {End-to-end architecture in cross-lingual ASR systems}
\label{fig:cross_architecture}
 
\end{figure}
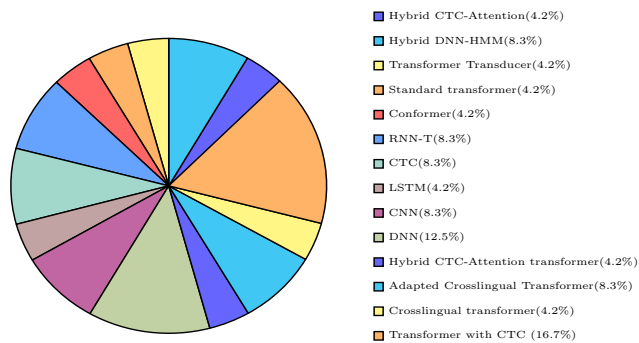


 
\subsection{Architecture-model and Decoding}
Cross-lingual ASRs are used in transformer architecture in 52.2\% of the research work, and 47.8\% of models are used in deep learning architectures. From the anysis depicted in Fig. \ref{fig:cross_architecture}, the deep learning model includes hybrid CTC-Attention \cite{r52}, hybrid DNN-HMM \cite{klejch2021deciphering} \cite{feng2021phonotactics}, RNN-T \cite{r51, r108}, CTC \cite{r127}, CNN-LSTM \cite{r50, fukuda2021knowledge}, DNN \cite{r49} while the transformers includes transformer transducer \cite{silovsky2023cross}, standard transformer \cite{wiesner2022injecting}, conformer \cite{ashihara2023exploration}, hybrid CTC-attention transformer \cite{r52}, adapted crosslingual transformer \cite{r53, r116}, crosslingual transformer \cite{r103}, transformer with CTC \cite{farooq2023learning, xue2023tranusr}. The decoding algorithm integrates the outputs of the acoustic model and language model to determine the most likely transcription for the input audio. Several algorithms are commonly used for decoding in cross-lingual ASR, such as greedy search, beam search, Viterbi and CTC. Beam search is the most used decoding approach \cite{r52, r51, r116}. It is a complex decoding algorithm that examines several options concurrently. It keeps a beam or group of candidates and selects the most likely candidates based on the total scores of the acoustic and language models. After the beam search, a greedy search was used. In greedy search, the most suitable phoneme, character, or word is chosen at each time step from the transcript, which makes it simpler \cite{polak2021coarse, likhomanenko2023unsupervised}, \cite{farooq2023learning}. Viterbi decoding is used in HMM-based systems, but in the end-to-end cross-lingual models, it is combined with CTC to perform decoding operations \cite{ashihara2023exploration}.

\begin{table*}
\scriptsize

\begin{tabular}[c]{p{1.5cm}p{1cm}p{5cm}p{5cm}}
\hline
\textbf{Reference} &
\textbf{Noisy} &
\textbf{Acoustic Features} &
\textbf{Architecture}\\
 \hline
  \cite{r50} &
 \ding{55} &
  39D MFCC  &
  CNN, LSTM \\
  \cite{r49} &
 \ding{55} &
 43D MFCC &
 DNN\\ 
  \cite{r51} &
\checkmark &
 80D MFCC  &
 RNN-T \\ 
  \cite{r105} &
 \ding{55} &
 40D Filterbank &
 DNN 
 \\ 
  \cite{r103} &
 \ding{55} &
  Bottleneck  &
  Cross-lingual transformer  \\ 
  \cite{conneau2020unsupervised} &
 \ding{55} &
 NA &
 Standard Transformer 
   \\ 
  \cite{feng2020unsupervised} &
  \checkmark &
  13-D MFCC  &
  DNN\\ 
  \cite{klejch2021deciphering} &
 NA  &
   
  40D MFCC  &
  Hybrid HMM-DNN \\ 
  \cite{feng2021phonotactics} &
  NA &
3D pitch , 40D MFCC, 100D ivectors  &
Hybrid HMM-DNN  \\ 
  \cite{r127} &
 \ding{55} &
  39D MFCC &
  CTC \\
  \cite{r52} &
 \ding{55} &
  80D filterbank, pitch  &
   Hybrid CTC-Attention transformer \\ 
  \cite{r53} &
 \ding{55} &
  80D MFCC,Pitch &
   Adapted crosslingual transformer \\ 
  \cite{fukuda2021knowledge} &
  \checkmark &
40D log Mel-frequency  &
CNN \\ 
  \cite{r116} &
 \ding{55}   &
  Convolutional &
  Adapted crosslingual transformer\\ 
  \cite{r101} &
 \ding{55}   &
Filterbank, 3D-Pitch, 40D-MFCC, 100D-i-vector &
Transformers with CTC \\ 
  \cite{ashihara2023exploration} &
   \ding{55} &
  80D FBANK  &
  Conformer \\ 
  \cite{r108} &
  \ding{55}  &
80D log-mel filterbank &
RNN-T 
\\ 
  \cite{wiesner2022injecting} &
 \ding{55} &
  convolutional &
  Standard transformer 
   \\ 
  \cite{likhomanenko2023unsupervised} &
 \checkmark & 
 80D log-mel filterbank&
 Transformers with CTC 
  \\
  \cite{farooq2023learning} &
   \ding{55} &
  40D filterbanks &
   Hybrid CTC-Attention \\ 

   \cite{xue2023tranusr} &
 \ding{55} &
 NA &
Transformer with CTC \\ 

\cite{hu2024cam} & \ding{55} & NA & CAM \\

\cite{chen2024improving} & \checkmark & 43D MFCC & TMPL, Transformers with CTC \\

   \hline
\end{tabular}
\caption{Acoustic features used for training model in the architecture of Crosslingual ASR systems.}
\label{tab:cross_acoustic}
 \end{table*}

\subsection{Evaluation Metric}

WER is the most commonly used metric, employed in about 70\% of cross-lingual ASR research studies (referenced in works like \cite{r51, conneau2020unsupervised, feng2020unsupervised, klejch2021deciphering, fukuda2021knowledge}). Works like \cite{xue2023tranusr, r103, conneau2020unsupervised} have leveraged PER, particularly in studies where phonetic precision is crucial as it accurately transcribes phonetic elements, especially in multilingual contexts where phonetic structures vary significantly. While less commonly used than WER and PER, CER has gained increasing importance in recent research, as in \cite{farooq2023learning}.

\begin{figure*}[h!]
\centering
\begin{tikzpicture}

  \begin{axis}[
    width=14cm,
    height=5cm,
    ybar,
    bar width=0.2cm,
    ylabel={\# Research Studies},
    ymin=0,
    ymax=14,
    xtick=data,
    legend pos=north east,
    symbolic x coords={MFCC, Filterbank, Pitch, i-vector, Mel-spectogram, Log-mel features},
    ytick={0,2,4,6,8,10,12,14}
    ]
    
    \addplot coordinates {(MFCC,4) (Filterbank,13) (Pitch,0) (i-vector,0) (Mel-spectogram,0)};
    \addplot coordinates {(MFCC,10) (Filterbank,8) (Pitch,3) (i-vector,4) (Mel-spectogram,0)};
    \addplot coordinates {(MFCC,9) (Filterbank,6) (Pitch,2) (i-vector,1) (Mel-spectogram,5)};
    \addplot coordinates {(MFCC,11) (Filterbank,6) (Pitch,5) (i-vector,2) (Mel-spectogram,0)};
    
    \legend{Monolingual, Bilingual, Multilingual, Crosslingual}
    
  \end{axis}
  
\end{tikzpicture}

\caption{The analysis gives an overview of acoustic features used by research work in monolingual, bilingual, multilingual, and cross-lingual speech recognition.}
\label{fig:acoustic_features_methodology}

\end{figure*}
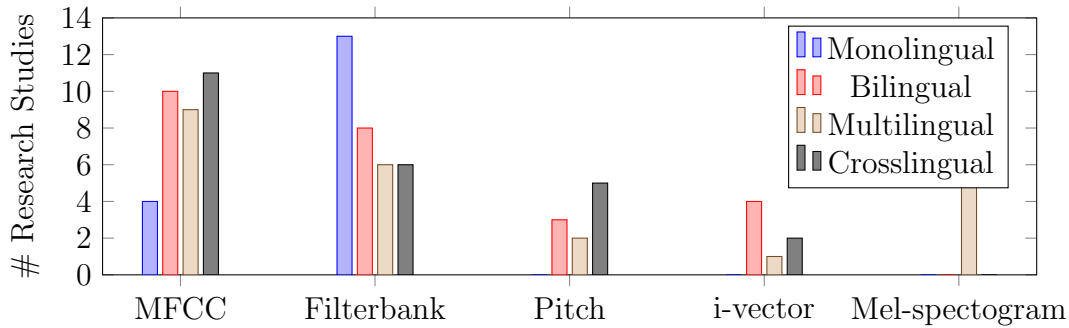

\section{ Popular Speech Transformer for Automatic Speech Recognition}

Lingvo \cite{shen2019lingvo} and the NeMo toolkit \cite{kuchaiev2019nemo} have gained popularity for large-scale ASR development due to their modular design and seamless integration with TensorFlow or PyTorch. Lingvo's modular approach is key for assembling complex neural networks, while NeMo’s focus on mixed-precision training reduces computational overhead (refer for detailed framework\cite{shen2019lingvo},\cite{kuchaiev2018mixed}).
We summarize our key findings in the Table \ref{tab:toolkits_for_asr}. Other significant ASR models and toolkits include Fairseq \cite{ott2019fairseq} for efficient training in neural network architectures, Linformer \cite{wang2020linformer} which restructures self-attention for linear complexity, Sync-Transformer\cite{tian2020synchronous} for synchronizing encoder and decoder in real-time ASR applications, Bi-encoder Transformer \cite{lu2020bi} designed for Mandarin-English code-switching, Wav2Vec 2.0 \cite{baevski2020wav2vec} and its extensions, which revolutionize ASR through self-supervised learning, SpeechT5\cite{ao2021speecht5} a unified-modal encoder-decoder framework, HuBERT \cite{hsu2021hubert} and UniSpeech\cite{wang2021unispeech} employing self-supervised methods, CTC-Transformer \cite{huang2021improving} for code-switching speech recognition, SpeechBrain\cite{ravanelli2021speechbrain}, a comprehensive, open-source speech processing toolkit, SpeechFormer \cite{chen2022speechformer} and Squeezeformer \cite{kim2022squeezeformer} for efficient ASR, Paraformer \cite{gao2022paraformer} and Branchformer \cite{peng2022branchformer} introducing novel architectures, Whisper \cite{radford2023robust} tailored for ASR in challenging environments, LA-NAT\cite{lin2023lexical} and Sumformer \cite{parcollet2023sumformer} enhancing accuracy and efficiency. LFEformer \cite{wei2023lfeformer} for improved local feature extraction, HyperConformer \cite{mai2023hyperconformer} combining Conformers and HyperMixers for efficient processing.
These toolkits focus on overcoming challenges like computational demands, code-switching complexity, and the need for models that cater to low-resource languages. 
\begin{table}[!htb]
\scriptsize

\begin{tabular}{p{4cm} p{1.5cm} p{1.5cm} p{1.5cm} p{1.5cm} p{1.2cm}} 

\label{tab:toolkit_table}\\
\hline

\textbf{Model} &
 \textbf{Parameters} &
   
 \textbf{Monolingual} &
  \textbf{Bilingual} &
 \textbf{Multilingual} &
  \textbf{Crosslingual} \\ \hline

ESPnet \cite{watanabe2018espnet} & 
Na &
 \checkmark &
\checkmark &
\checkmark &    
\checkmark \\ 

Lingvo \cite{shen2019lingvo}& 
NA &
 \checkmark &
\checkmark  &
 &    
\checkmark \\
 
 Nemo \cite{kuchaiev2019nemo} & 
NA &
 &
 &
 &    
\checkmark \\

Fairseq \cite{ott2019fairseq} & 
NA &
 \checkmark &
\checkmark &
\checkmark &    
\checkmark \\ 

 Speech2Text \cite{wang2020fairseq} & 
  263M &
 &
 &
\checkmark &
   \\ 
Linformer  \cite{wang2020linformer} & 
  NA &
\checkmark &
&
 &
   \\ 
Sync-Transformer \cite{tian2020synchronous} & 
 NA  &
\checkmark &
&
 &
   \\


Wav2Vec2.0 \cite{baevski2020wav2vec} & 
  95M &
\checkmark &
 &
 &
   \\
Wav2Vec2-Conformer \cite{r28} & 
  600M, 1B &
\checkmark &
 &
 &
   \\ 
XLSR-Wav2Vec2 \cite{conneau2020unsupervised}& 
  NA &
&
 &
&
 \checkmark \\  
 Biencoder Transformer \cite{lu2020bi} & 
   45.6M &
 &
\checkmark &
 &
   \\ 
Hubert \cite{hsu2021hubert} & 
  1B &
\checkmark &
&
 &
   \\ 
SpeechT5 \cite{ao2021speecht5} & 
 NA &
\checkmark &
 &
 &
   \\ 
UniSpeech \cite{wang2021unispeech} & 
  NA &
 &
 &
 &
  \checkmark \\

Speech Brain \cite{ravanelli2021speechbrain} & 
NA &
 \checkmark &
 &
\checkmark &    
 \\

CTC-Transformer \cite{huang2021improving} & 
  NA &
 &
\checkmark &
 &
   \\
Transformer\-Transducer \cite{dalmia2021transformer} & 
  NA &
 &
  \checkmark &
 &
   \\ 
XLS-R \cite{babu2021xls} & 
  2B &
&
 &
 &
  \checkmark \\ 
SEW \cite{wu2022performance} & 
  177M &
\checkmark &
 &
 &
   \\ 
UniSpeech-SAT \cite{chen2022unispeech} & 
  316.61M &
\checkmark &
 &
 &
   \\ 
WavLM  \cite{chen2022wavlm} & 
  94.70M &
\checkmark &
 &
&
   \\ 
Squeezeformer  \cite{kim2022squeezeformer} & 
   236.3 M &
\checkmark &
 &
 &   
   \\ 
Paraformer  \cite{gao2022paraformer} & 
  63M &
\checkmark &
 &
 &
   \\ 
Branchformer \cite{peng2022branchformer} & 
  45.4M &
\checkmark &
 &
 &
   \\ 
Speechformer \cite{chen2022speechformer} & 
NA &
\checkmark &
&
\\  
E-Branchformer \cite{kim2023branchformer} & 
  148.9M &
\checkmark &
 &
&
   \\ 
LFEformer \cite{wei2023lfeformer} & 
  30.68M &
\checkmark &
&
 &
   \\ 
LA-NAT \cite{lin2023lexical} & 
  79.8 M &
\checkmark &
 &
 &
   \\   
Whisper \cite{radford2023robust} & 
  1550M &
&
 &
\checkmark &
   \\ 
MMS \cite{pratap2023scaling} & 
  1B &
 &
 &
\checkmark &
   \\ 
HyperConformer \cite{mai2023hyperconformer} & 
  93.3M &
\checkmark &
 &
 &
   \\ 
Sumformer  \cite{parcollet2023sumformer} & 
  65M &
 &
 &
\checkmark &
   \\ 
Transformer\-Transducer \cite{yu2023code} & 
  NA &
 &
\checkmark &
&
   \\

SpeechFormer++ \cite{chen2023speechformer++} & 
 66.79M &
 \checkmark &
 &
&    \\  

CAM \cite{hu2024cam} & 9.4M & & & \checkmark \\

Whistle \cite{yusuyin2024whistle} & 90M & & \checkmark & \checkmark \\

\hline
\end{tabular}
\caption{State-of-art (SOTA) toolkits/framework/models/libraries transformer model for ASR along with their parameters and linguistic capabilities}
\label{tab:toolkits_for_asr}
\end{table}

\section{Discussion: Open Challenges and Future Research Directions}
The promising architecture of deep learning frameworks and transformers significantly impacts the advancement of speech or audio processing. We particularly focus on various levels of lingualism, especially for low-resource language requiring a multi-facet approach. To the best of our knowledge, we cover extensively all the state-of-art research based on the monolingual in section \ref{sec:mono}, bi-lingual in section \ref{sec:bi}, multilingual in section \ref{sec:multi} and cross-lingual in section \ref{sec:cross}. However, we still believe in the need for resources or scarcity of resources to build ASR for low-resource languages. This linguality perspective, explored in this paper, faces many key challenges and opens the avenue for further research in this domain.  

\begin{table*} 
\vspace{-2ex}
\tiny
\begin{tabular}{p{2cm}p{4.5cm}p{3.5cm}p{0.4cm}p{0.4cm}p{0.4cm}p{0.4cm}p{0.4cm}}
\hline
 \textbf{Dataset} &\textbf{Languages} &\textbf{Duration}  & \textbf{Mono} & \textbf{Bi} & \textbf{Multi} & \textbf{Cross} & \textbf{Open source}\\ \hline

Librispeech \cite{panayotov2015librispeech} & 
EN &
960h &
\checkmark &
\checkmark &
\checkmark &
\checkmark &
\checkmark \\
Google voice traffic &
EN &
18,000h &
&&&&  \ding{55}\\
WSJ \cite{paul1992design} &
EN &
81h &
&
&
\checkmark &
& 
\checkmark\\
Microsoft’s Cortana voice assistant system  \cite{chen2021developing}& 
EN &
3400 h &
\checkmark &
\checkmark &
&
&
 \ding{55}\\
Chime &
EN &
342h &
\checkmark &
&
\checkmark
&
&
\checkmark \\
RATS \cite{graff2014rats}  &
EN, FA, AR, PS, URD, EN, PRS &
402h&
&
&
&
&
 \ding{55}\\
AISHELL-1 \cite{bu2017aishell} &
ZH &
170 &
\checkmark &
\checkmark &
\checkmark &
&
\checkmark  \\
AISHELL-2 \cite{du2018aishell} &
ZH &
1000h &
\checkmark &
\checkmark &
&
&
\checkmark \\
TEDLIUM \cite{rousseau2014enhancing}&
EN &
452h &
&
&
\checkmark &
&
\checkmark \\
Corpus of Spontaneous Japanese (CSJ) \cite{maekawa2000spontaneous} &
JA &
1000h &
&
&
\checkmark &
\checkmark &
\checkmark \\
SEAME \cite{lyu2015mandarin} &
ZH-EN &
30h &
&
\checkmark &
&
&
\checkmark \\
FAME \cite{yilmaz2016longitudinal} &
FRR-NL &
18.5h &
&
\checkmark &
&
&
\checkmark \\
ASRU 2019 \cite{shi2020asru} &
ZH-EN &
740h &
&
\checkmark &
&
&
 \ding{55} \\
Microsoft live Cortana	&
ZH-EN &
ZH: 4000h, EN: 3400h, CS: 300h &
&
\checkmark &
&
&
  \ding{55}\\
IARPA BABEL \footnote{https://www.ldc.upenn.edu/search/node/babel} &
ZH, TL, PS, VI, AS, BN, HT, LO, TA, ZU, SW, KU, TPI, CEB, KAZ, TE, LT, GN, IG, AM, MN, JV, LUO &
(213, 201, 205, 215, 203, 207, 350, 211, 350, 203, 200, 203, 201, 210, 198, 207, 204, 204, 204, 204)h &
&
&
\checkmark &
\checkmark &
\checkmark \\

Euronews corpus \cite{roberto2014euronews} &
AR, EN, FR, DE, IT, PL, PT, RU, ES, TR	&
 940h &
&
&
\checkmark &
&
\checkmark \\
Commonvoice	\cite{ardila2019common} &
AB, AR, EU, BR, CA, ZH, CV, DV, NL, EN, EO, ET, FR, DE, CNH, Chin, ID, 	IA, GA, IT, JA, KAB, RW, KY, LV, MN, FA, PT, RU, Sakha, SL, ES, SV, TA, TT, TR, VOT, CY  &
(<1, 15, 83, 10, 120, 12, 43, 2, 8, 23, 1087, 16, 12, 184, 340, 4, 5, 2, 3, 40, 2, 192, 20, 8, 9, 70, 30, 31, 6, 5, 31, 3, 5, 26, 10, <1, 48)h &
\checkmark &
\checkmark &
\checkmark &
\checkmark &
\checkmark \\
VoxPopuli \cite{wang2021voxpopuli} &
EN, DE, FR, ES, PO,  IT, RO, HU, CS, NL, FI, HR, SK, SL, ET, LT, PT, BG, EL, LV, MT, SV, DA &
(543, 282, 211, 166, 111, 91, 89, 63, 62, 53, 27, 43, 35, 10, 3, 2)h &
\checkmark &
&   
&
\checkmark &
\checkmark \\

GlobalPhone	 \cite{schultz2013globalphone}& 
AR, BG, HR, CS, FR, DE, HA, JA, KO, ZH, PT, PL, RU, SUJI, ES, SV, TA, TH, TR, UK, VI &
(12, 16.47, 11.48, 26.49, 24.55, 14.54, 6.36, 21.51, 16.34, 26.38, 22.45, 18.39, 21.08, 9.5, 17.35, 17.39, 15.5, 19.05, 13.04, 11.32, 22.15)h  &
&
&
\checkmark &
\checkmark &
\checkmark \\

Multilingual Librispeech \cite{pratap2020mls} &
EN, DE, NL, FR, ES, IT, PT, PO &
(1,966.51, 1,554.24, 1,076.58, 917.68, 247.38, 160.96, 103.65)h &
&
\checkmark &
\checkmark &
\checkmark &
\checkmark \\
Ethiopian languages corpora &
AM, TI, OM, WAL &
(24, 22.1, 22.8, 29.7)h &
&
&
&
&
\\
SpeechOcean \cite{zhang2021speechocean762} &
GU, TE, TA &
(50, 50, 50)h &
&
&
\checkmark &
&
\checkmark\\
Google's voice search traffic &
EN(US), EN(IN), ES(US), PT, ES, AR(GULF), AR(EG), HI(IN), MR (IN), BN, ZH, RU, TUR, HU, MS &

(53.5K, 27.1K, 47.6K, 32.9K, 23.5K, 11.9K, 11.9K, 32.3K, 16.7K, 16.5K, 22.8K, 22.8K, 22.1K, 9.9K, 7.6K)h &
&
&
\checkmark &
&
 \ding{55}\\

Google indic corpus \cite{he2020open} &
GU, MR, KN, ML, TA, TE &
(7.89, 3.02, 4.48, 5.51, 7.08, 5.71)h &
&
&
 &
\checkmark &
 \ding{55} \\
IIT Madras ASR Challenge \footnote{https://github.com/Speech-Lab-IITM/Hindi-ASR-Challenge} &
HI, TA &
(50, 108)h &
&
&
&
\checkmark &
\checkmark \\
Finnish parliament dataset \cite{jain2020finnish} &
FI &
1500h &
\checkmark &
&
&
&
 \ding{55}\\
MALACH \cite{ramabhadran2003towards} &
EN, CS &
200h &
\checkmark &
&
&
&
 \ding{55} \\
Gigaspeech &
EN&
10000h &
\checkmark &
&
&
&
\checkmark \\ 
African Accented French \footnote{https://www.openslr.org/57} &
FR&
22h &
\checkmark &
&
&
&
\checkmark \\
Canadian French Emotional Speech \cite{gournay2018canadian} &
FR &
936 samples &
\checkmark &
&
&
&
\checkmark \\
HKUST \footnote{https://catalog.ldc.upenn.edu/LDC2005S15} &
ZH &
149h &
\checkmark &
&
&
&
 \ding{55}\\
Switchboard-1 \footnote{https://catalog.ldc.upenn.edu/LDC2005S15} &
ZH &
260h &
\checkmark &
\checkmark &
\checkmark &
&
 \ding{55}\\
Fisher \footnote{https://catalog.ldc.upenn.edu/LDC2004T19} &
EN &
984 h &
\checkmark &
\checkmark &
&
\checkmark &
 \ding{55}\\
Artie Bias Corpus \cite{meyer2020artie} &
EN &
2.4h &
&
&
\checkmark &
&
\checkmark \\
NCHLT Speech Corpus \cite{barnard2014nchlt} &
AF, NBL, XH, ZU, NSO, ST, TSN, SS, VEN, TS &
800h&
&
&
\checkmark &
&
 \ding{55} \\
Nordics \cite{gaur2022multilingual}&
DA, FI, NO, SV, NL, EN &
(3.3k, 3.8k, 5.4k, 7.5k, 8.4k, 52k)h &
&
&
\checkmark  &
&
 \ding{55} \\
LJ Speech \footnote{https://keithito.com/LJ-Speech-Dataset/} &
EN &
23:55:17h &
&
&
\checkmark &
&
\checkmark \\
FLEURS \cite{conneau2023fleurs} &
102 languages &
492k &
&
&
\checkmark &
&
 \ding{55}\\
\hline
 
\end{tabular}
\caption{The popular corpus used for building/training ASR}
\label{tab:dataset_list_for_asr}
\end{table*}

\textbf{Large scale requirement of data or handling low-resource languages} In ASR, the emergence of transformer-based architectures has highlighted a pivotal challenge: their significant dependence on large volumes of training data. \cite{mehmood2022fednst} indicate that the performance of ASR models is proportionate to the training corpus size. This correlation is particularly evident across diverse lingualism spectrum's, including monolingual, bilingual, multilingual, and cross-lingual systems. However, the availability of such large-scale datasets is predominantly limited to widely-spoken languages, such as English and Mandarin, leaving a gap in resources for low-resource languages. Collecting and annotating substantial speech data for these languages present significant challenges\cite {maison2023improving}. This challenge could be mitigated by transcribing audio across multiple languages, \cite{conneau2020unsupervised}, which seems much more challenging than compiling a monolingual dataset. To mitigate these challenges, the most prominent solution is the generation of synthetic data from existing datasets using advanced data augmentation techniques \cite{zhang2021xlst}, including speech perturbation, volume perturbation, audio-splicing, noise injection, and pitch shifting, which are key techniques for increasing the diversity and size of the training corpus.
Another direction could be the compilation of multilingual datasets from various sources, followed by model training using weak supervision methods \cite{radford2023robust}. These techniques enable the efficient utilization of available data while addressing the scarcity of large annotated datasets. Transfer learning is another strategic approach where models pre-trained on extensive datasets are adapted to smaller, specific datasets. \cite{likhomanenko2023unsupervised}, demonstrating its effectiveness in enhancing the performance of models tailored for low-resource languages. Moreover, applying multitask learning paradigms, where a model is trained concurrently on multiple tasks, can significantly enrich its linguistic capabilities and adaptability. Integrating semi-supervised and unsupervised learning methodologies, where transcription of input data is not mandatory, offers a path forward in scenarios of limited annotated data availability. These approaches, as discussed by\cite{farooq2023learning}, are gaining traction as viable alternatives to traditional supervised learning methods in ASR. Table \ref{tab:dataset_list_for_asr} represents a list of different datasets, languages and their size along with the lingualism approaches, it has been explored.  Future research in ASR, particularly concerning low-resource languages, requires the benchmark for dataset volume required for efficient model training. Such benchmarks, as proposed by\cite{silovsky2023cross}, would provide a framework for developing more targeted and effective ASR systems, facilitating progress in this field and linguistic resources and technology.

\textbf{Computational challenge, efficiency and its optimization strategy}, The application of large-language models or transformer-based end-to-end models\cite{vaswani2017attention} to speech recognition system is elaborated in Section \ref{sec:mono}, \ref{sec:bi}, \ref{sec:multi} and \ref{sec:cross}, have gained significant success but also highlight the computational challenge. This rise in computational demands, compared to RNN  training, requires the development of optimizers and learning rate schedulers. Specifically, the self-attention mechanism applied to speech poses quadratic computational complexity and does not convey specific meaning compared to words or lexical units. Rather, it complicates the computation of attentive weights. To address and enhance the self-attention mechanism for speech, \cite{yeh2019transformer} adopts a pre-processing mechanism such as convolutions that group adjacent speech frames into meaningful units (phonemes). For longer speech sequences \cite{dong2019self} propose time-restricted or truncated self-attention with\textit{ down sampling and pooling the audio}. The role of { positional encoding} in transformers, crucial for capturing longer dependencies in sequential data, also poses challenges. The original sinusoidal position encoding can hinder performance in speech-based systems due to the extended length of sequences and may exhibit poor generalization in certain conditions. Various alternative approaches have been explored to address these issues in ASR, though further exploration is needed in other speech-related domains.
Their high computational cost contrasts the success of transformer-based end-to-end models in speech-related areas during \textit{inference}, especially when compared to LSTM models \cite{lu2020exploring}. The Transformer Transducer (T-T) model's computational cost increases significantly with input sequence length, limiting its practical use. The subsequent development of the Conformer Transducer (C-T) aimed to improve T-T. Still, its non-streamability due to full-sequence attention in its encoder remains a limitation for streaming applications \cite{chen2021developing}. Chen et al. demonstrated the effectiveness of s\textit{treaming processing and early stopping} in reducing latency and runtime costs in speech models \cite{chen2021developing}. Furthermore, Lin et al. explored weight pruning, head pruning, low-rank approximation, and knowledge distillation to reduce model parameters \cite{lin2022compressing}. However, optimizing transformer models for different hardware platforms can introduce challenges like load imbalance, communication overhead, or memory fragmentation. Techniques for improving hardware utilization include tensor decomposition, kernel fusion, mixed precision arithmetic, and hardware-aware optimization
\cite{gu2022heat}. Addressing the self-attention mechanism's quadratic complexity concerning sequence length has led to several proposed solutions, including sparse attention patterns, low-rank factorization, random feature maps, and locality-sensitive hashing. Additionally, transformers' memory consumption, scaling linearly with sequence length and quadratically with hidden dimension size, poses significant challenges for large-scale data training and inference. Solutions such as reversible residual connections, gradient checkpointing, weight sharing, and parameter pruning have been suggested to manage memory usage.
Recent research has been directed toward optimizing transformer architectures across diverse lingual contexts to address these computational inefficiencies. \cite{kim2023branchformer}, explored local-global context merging, and \cite{jain2020finnish}, focusing on reducing architectural parameters at the encoder or decoder levels which would increase the performance with lesser computational load. In monolingual ASR, methodologies such as downsampling audio signals have been investigated \cite{andrusenko2023uconv}, aiming to streamline the processing pipeline. Optimization efforts have been primarily concentrated at the decoding level for bilingual ASR systems to enhance efficiency without compromising the linguistic nuances \cite{lin2023lexical}. Recent research has focused on architectural modifications for multilingual and cross-lingual models. This includes transitioning the convolutional layer structure to a more efficient 1-D format \cite{winata2020lightweight} and implementing Non-Autoregressive (NAR) models, which have shown potential in reducing training time \cite{lin2023lexical}.
Additionally, byte-level input processing and optimizing batch sizes can be explored to enhance model performance \cite{soky2023domain}. Enhancing the efficiency of Transformers remains a key concern due to their large and complex architectures. Efforts to optimize efficiency involve using fewer training data and parameters, including strategies like knowledge distillation, model compression, asymmetrical network structures, and improved training sample utilization. Significant contributions in this regard include knowledge distillation, model simplification, weight-sharing, training with fewer samples, and compressing and pruning the model. These efforts highlight the ongoing importance of addressing efficiency in Transformer development. One of the possible solutions could be transfer learning. By leveraging models pre-trained on large datasets, researchers have efficiently fine-tuned these models for specific tasks, substantially reducing the training duration compared to training from scratch \cite{r53}. The knowledge captured in pre-trained models adapts to specific lingual contexts, demonstrating resource-efficient methodology for ASR model training. The necessity for substantial memory to store large-scale datasets and the model's dense network parameters remains a substantial technical challenge. For future research, more work on the architectural level for model and efficient data storage and processing techniques could be explored to process low-resource and endangered language environments. 

\textbf{Robustness challenge: handling noisy data and domain mismatch}
The ASR system trained on large-scale corpus has achieved impressive performance, surpassing the current state-of-art across various lingualism frameworks and its wide adoption to \textit{down-streaming task}. However, the robustness of these models, especially in \textit{noisy data environment} and \textit{domain shifts }, is a key challenge\cite{r125, r96, r93}. 
For addressing issues related to noisy or distorted speech signals, \textit{audio-visual ASR}, which leverages auditory and visual cues (such as lip movements) to improve recognition accuracy and robustness against background noise, could be a promising approach. It provides a novel direction for enhancing ASR system resilience; even conformer architecture could be utilized for this approach. Another crucial aspect adversely affecting ASR model performance is its dependence on \textit{raw speech features}, which makes them more exposed to noise inference and speaker variation. The lack of modelling of prosodic features such as rhythm, stress, and intonation in speech data impacts the performance. These variations are also caused by \textit{speakers and speaking style}, which calls for an unsupervised approach that learns speaker embeddings from raw speech using\textit{ multi-task learning framework}, where speaker recognition and ASR task both are jointly optimized with contrastive loss. In the future, it calls for developing a robust and adaptable speech recognition model to address these challenges for diverse acoustic conditions, speaker variations, and a wide range of linguistic variations. 
Models trained in a specific domain frequently perform poorly when confronted with audio from different domains. Addressing this \textit{domain mismatch} challenge involves advanced\textit{ domain adaptation techniques}, including unsupervised and semi-supervised learning methods.  To overcome this, \cite{thomas2022efficient} have demonstrated significant progress, showcasing improvements over traditional supervised training methods. These models trained on mono-lingual face difficulty in adapting to other domains and generalize to another language, which degrades the performance. It also leads to the common issue in the ASR system of handling \textit{out-of-vocabulary (OOV) words}, which often emerge from domain-specific terms, proper nouns, slang, or neologisms. To overcome this,\cite{jain2023phonological} uses contrastive learning and focuses on training models using multiple languages with aligned linguistic features to enhance generalization in bilingual, cross-lingual, and multilingual ASR. The self-supervised technique enables the models to differentiate between similar and dissimilar speech segments. By using a multilingual phonetic vocabulary, these models can capture cross-lingual similarities, making transfer learning across different languages. Additionally, as we mentioned in the first challenge of large-scale data requirements, self-training and semi-supervised learning methods are used for unlabeled data, which enhances the quality of speech representations and would further improve performance by achieving state-of-the-art results for low-resource languages. Further research to enhance model adaptability to diverse acoustic and linguistic environments or real-world scenarios would bridge the gap in the research and usability of ASR applications.

\textbf{Generalization and Transferability Challenge}
One of the most prominent challenges of current ASR is its limitation to adapt to \textit{new languages} or \textit{domains} across a range of tasks and scenarios for diverse applications. This issue arises due to their dependence on extensive data for generalization, which might be biased. The dependence impacts the performance in downstream ASR, especially if the training data lacks a diverse set of examples or quality. To address this, \cite{xue2021bayesian} introduces the Bayesian Transformer Language Model (BTLM), which integrates a Bayesian framework to improve the model’s adaptability to out-of-domain speech data. The Bayesian Transformer employs variational inference to estimate latent parameter posterior distributions, which can effectively manage model uncertainty. \cite{lin2022compressing} applies compression techniques such as weight pruning, head pruning, low-rank approximation, and knowledge distillation to transformer models. This strategy, especially with contrastive predictive coding (CPC), has shown enhanced performance in handling out-of-domain ASR tasks compared to traditional language models. \cite{zhou2019improving} proposed Parallel Scheduled Sampling (PSS) and Relative Positional Embedding (RPE), which enhances robustness and mitigates exposure to bias by sampling tokens from either the ground truth or predicted sequences during training. By encoding relative distances between tokens, RPE improves the transformer's capacity for processing long speech sequences. Despite the challenge of adapting to a new language, it is more difficult to keep up the pace of ever \textit{evolving languages, slang} due to the dynamic nature of language. New slang, terms, and colloquialisms frequently emerge, which calls for \textit{dynamic language models} that can quickly adapt to evolving language use, perhaps through continuous online learning mechanisms. 
The transferability across different domains poses another challenge. To mitigate data augmentation, adversarial perturbation strategies, learning shared embedding spaces, and knowledge distillation could be utilized. In the case of multimodal ASR, there are distribution gaps between training and practical data, which induces difficulty in transferability. \cite{zhan2021product1m} presents the difficulty in transferring supervised multimodal transformers trained on well-aligned cross-modal pairs to weakly aligned test data. Further, adapting them to different tasks and different languages requires fine-tuning. It requires continual learning to enhance the versatility and efficacy in diverse linguistic and domain generalization and transferability.

\textbf{Multimodal Learning} 
In ASR, capturing the nuances of speech beyond audio, such as visual cues from lip movement or facial expressions, potentially improves recognition accuracy in complex environments but also induces fusion challenges. The fusion strategy is prominently categorized into three levels: early, middle, and late fusion. Early fusion occurs at the input stage, combining different modalities before processing. Middle fusion involves merging intermediate representations of different modalities within the transformer's attention mechanism \cite{sahay2020low}. This process might include direct integration of bimodal representations or more complex schemes such as alternating attention mechanisms across modalities. Late fusion, less common in MML ASR, focuses on combining the outcomes of separate modal pathways at the final prediction stage. Inspired by BERT's success, early fusion in one-stream architectures requires handling modality-specific inputs with varied masking techniques. It poses a significant challenge in effectively processing distinct input types while maintaining the integrity of each modality's information. Middle fusion, while offering more flexibility in combining modality-specific features at different network depths, demands sophisticated strategies to balance and adaptively combine these features \cite{sahay2020low}. Another aspect of ASR is managing the intrinsic synchronization between modalities, such as audio and visual data, in audio-visual speech recognition. The challenge lies in aligning these modalities within the transformer framework to leverage the complementary information effectively. Apart from the technical challenges, they exhibit computational overheads, to overcome techniques such as bottlenecks token-based fusion aims to streamline this process but still requires careful consideration of computational resources. Additionally, optimizing these models for efficient training and inference, given the increased data and parameter load, remains a challenge.
As mentioned in the previous challenge, adapting these multimodal transformers to new domains or tasks within ASR, such as adapting a model trained on telephony speech to process audio from video conferencing, involves overcoming domain gaps. Techniques like data augmentation and adversarial training have been proposed to enhance model transferability, but effective implementation in ASR requires careful calibration to preserve speech-specific characteristics. Multiple research works use large-scale crawled web corpora or techniques like contrastive learning for cross-modal alignment in ASR, which introduces additional challenges. These include ensuring synchronization between audio and corresponding text or visual data and adapting these models for ASR-specific tasks.

\textbf{Language mismatch and code-mixing}
Cross-lingual ASR can process speech across different source and target languages. However, it becomes challenging for diverse pronunciations specifically, in handling the disparities in phonetic inventories, pronunciation norms, and language-specific characteristics\cite{feng2021phonotactics} and further explored by \cite{feng2020unsupervised}. The problem grows with code-switching scenarios, where speakers alternate between languages within a single utterance.  Further research could be explored while some studies delve into the challenges of developing robust language identification algorithms for such language shift scenarios.
Additionally, managing vocabulary and lexicon mappings between the source and target languages is critical, and it includes addressing differences in word choices, synonyms, idiomatic expressions, and specialized terminologies unique to each language. Handling these linguistic variations requires models to interpret and map these lexical disparities accurately.
To overcome continual learning and adapt to linguistic divergence, some studies use deep learning to improve accuracy in code-switched speech recognition, and incorporating multilingual training data into these models efficiently works. However, further research in code-mixing on improved language identification, robustness to the diverse acoustic environment, data-scarce scenarios, adaptability, semantic understanding, contextual processing, multilingual input integrations, personalizing, and bias mitigation could be explored. 

\textbf{ Personalisation and speaker variability}
Personalizing ASR through user-specific adaptation or individuals with unique speech patterns, accents, or speech impairments is a key area in the ASR domain. On the other hand, speaker variability which varies in a wide range of acoustic differences including \textit{accent, speech rate, pronunciation, and speaking style} is the primary challenge in the performance degradation of ASR. Multiple research focuses on recognizing and transcribing speech considering these variations \cite{saon2023diagonal}. For bilingual ASR systems, the complexity is increased by the need to account for acoustic variations stemming from different languages, each with its unique\textit{ phonetic and prosodic} characteristics\cite{r75}. These systems must recognize speech in two languages and adapt to the variations in accents, dialects, and individual speaker idiosyncrasies within each language. In the case of multilingual ASR systems, the challenge extends to a broader range of linguistic characteristics. Each language brings distinct acoustic properties influenced by its unique accents and dialects, requiring the ASR system to be versatile and adaptive across diverse linguistic spectra. Cross-lingual ASR systems are the toughest to be capable of handling acoustic variations not only within a single language but also across languages. This involves accurately modelling and adapting to the differences in \textit{accents, dialects, and speaker characteristics} prevalent in both the source and target languages, ensuring the system's robustness in cross-lingual speech recognition \cite{r105}.
Further research on developing acoustic models that can adapt to speaker variability with diverse accents and dialects extract features from acoustic patterns. The complexity increases with non-standard speech patterns such as \textit{speech impairments, disfluencies and regional slang's} and prosodic features such as rhythm, stress, and intonation in speech, which can be crucial for understanding and disambiguating spoken language. A key research gap is exploring \textit{Paralinguistic features} specific to individual speaker characteristics such as pitch, volume, and speaking rate. This advancement also brings \textit{ethical implications and privacy concerns} due to the use of personal and potentially sensitive speech data. This calls for investigating secure and privacy-preserving ASR methods, such as federated learning and differential privacy, to protect user data while maintaining model efficacy. This challenge is enhanced by using energy-efficient ASR models that operate through mobile or IoT devices or interactive ASR systems that can learn from user feedback in real time, adjust and improve based on user corrections and preferences.

\section{Conclusion}
Transformers have demonstrated remarkable success in various artificial intelligence tasks, largely due to the recent prevalence of the self-attention mechanism. This mechanism has been pivotal in capturing long-term dependencies, yielding phenomenal outcomes in speech processing and recognition tasks. However, a significant limitation of these end-to-end architectures is their reliance on large-scale datasets, which poses challenges for advancing ASR development in low-resource languages. This paper presents a groundbreaking comprehensive survey of transformer applications, particularly from a lingualism perspective in audio and speech processing. We've observed an emerging trend toward adopting transformers for the development of ASR systems in low-resource languages. The main insights from our study include: (1)\textit{comparison of traditional and end-to-end ASR models}, as traditional ASR models require handcrafted features and often perform slowly due to their sequential nature. In contrast, recent end-to-end models use deep neural networks for transcribing audio into text, supporting parallel processing for significantly faster system performance. (2) \textit{advancements in foundational models} employ self-attention mechanisms to capture long-range dependencies, enabling training on hundreds to thousands of hours of data, including code-switching scenarios. Multilingual and cross-lingual models apply transfer learning techniques with these foundational models to assist low-resource languages by transferring knowledge from high-resource languages. (3) \textit{speech datasets and augmentation techniques}, the Librispeech corpus is prominently used for monolingual, the Mandarin-English SEAME corpus for bilingual, and the Common Voice for multilingual models. A review of different speech datasets detailing the number of languages, duration, and availability is depicted. Various augmentation techniques like SpecAugment, noise injection, and speed and length perturbation are applied across mono, bi, multi, and cross-lingual models to enhance robustness. (4) \textit{analysis of popular speech transformers}, their parameters, and their utilization in different lingualism techniques. (5) \textit{we discuss challenges and future directions} such as computational requirements, robustness, generalization, language mismatch, and multimodal learning. Future research should focus on enhancing cross-lingual/multilingual systems and addressing the performance challenges identified in this review. These findings and recommendations aim to guide researchers and developers in the speech-processing field, particularly emphasising expanding the capabilities and accessibility of ASR systems for diverse linguistic contexts. This survey underscores the transformative impact of transformers in ASR and highlights the path forward for addressing the unique challenges in this domain.

 \bibliographystyle{elsarticle-num}
 \bibliography{reference}





\end{document}